%% file: main.tex
\documentclass[]{fairmeta}

\usepackage{microtype}
\usepackage{graphicx}
\usepackage{subcaption}
\usepackage{csquotes}
\usepackage{afterpage}

\usepackage{amsmath}
\usepackage{amssymb}
\usepackage{mathtools}
\usepackage{amsthm}
\usepackage{xspace}
\usepackage{colortbl}
\usepackage{multirow}
\usepackage{enumitem}

\usepackage{nicefrac}

\usepackage{algorithm}
\usepackage{algorithmic}
\usepackage{amsmath}
\usepackage{amssymb} 
\usepackage{booktabs}
\usepackage{caption}
\usepackage{enumitem}
\usepackage{graphicx}
\usepackage{url}
\usepackage[table]{xcolor}
\usepackage{multirow}

\usepackage{setspace}
\usepackage{wrapfig}
\usepackage{tikz}
\usetikzlibrary{positioning}
\usepackage[most]{tcolorbox}

\usepackage{xcolor}

\makeatletter
\DeclareRobustCommand\onedot{\futurelet\@let@token\@onedot}
\def\@onedot{\ifx\@let@token.\else.\null\fi\xspace}

\def\eg{\emph{e.g}\onedot}

\def\etc{\emph{etc}\onedot} 
\def\vs{\emph{vs}\onedot}

\makeatother

\usepackage{orcidlink}

\definecolor{w_1}{RGB}{66,138,244}
\definecolor{w_2}{RGB}{73,144,245}
\definecolor{w_3}{RGB}{79,148,246}
\definecolor{w_4}{RGB}{87,154,247}
\definecolor{w_5}{RGB}{95,160,248}

\definecolor{uw}{RGB}{197,180,228}

\newcommand{\ours}{\textbf{\textsf{\textcolor{w_1}{U}\textcolor{w_2}{n}\textcolor{w_3}{i}\textcolor{w_4}{W}\textcolor{w_5}{M}}}}

\definecolor{fbApp}{HTML}{c8e7fa}
\definecolor{fbPurple3}{HTML}{f0ebf5}

\definecolor{citecolor}{HTML}{0071BC}
\definecolor{linkcolor}{HTML}{ED1C24}

\definecolor{citecolor}{HTML}{0071BC}
\definecolor{linkcolor}{HTML}{ED1C24}

\title{\emph{Towards \textsf{\textcolor{w_1}{\underline{U}}\textcolor{w_2}{\underline{n}}\textcolor{w_3}{\underline{i}}}fied \textsf{\textcolor{w_4}{\underline{W}}}orld \textsf{\textcolor{w_5}{\underline{M}}}odels for Visual Navigation via Memory-Augmented Planning and Foresight}}

\author[1,*]{Yifei Dong}
\author[1,*]{Fengyi Wu}
\author[1]{Guangyu Chen}
\author[2]{Lingdong Kong}
\author[1]{Qiyu Hu}
\author[1]{Yuxuan Zhou}
\author[1]{Xu Zhu}
\author[3]{Jingdong Sun}
\author[1]{Jun-Yan He}
\author[4]{Qi Dai}
\author[5]{Alexander G. Hauptmann}
\author[1,\dagger]{Zhi-Qi Cheng}

\affiliation[1]{University of Washington}
\affiliation[2]{National University of Singapore}
\affiliation[3]{Apple}
\affiliation[4]{Microsoft Research}
\affiliation[5]{Carnegie Mellon University}

\contribution[*]{Equal contribution, listed in random order}
\contribution[\dagger]{Corresponding author.}

\abstract{
    Enabling embodied agents to imagine future states is essential for robust and generalizable visual navigation. Yet, state-of-the-art systems typically rely on modular designs that decouple navigation planning from visual world modeling, which often induces state--action misalignment and weak adaptability in novel or dynamic scenarios. We propose \ours, a unified, memory-augmented world model that integrates egocentric visual foresight and planning within a single multimodal autoregressive backbone. UniWM explicitly grounds action selection in visually imagined outcomes, tightly aligning prediction with control. Meanwhile, a hierarchical memory mechanism fuses short-term perceptual cues with longer-term trajectory context, supporting stable and coherent reasoning over extended horizons. Extensive experiments on four challenging benchmarks (Go Stanford, ReCon, SCAND, HuRoN) and the 1X Humanoid Dataset show that UniWM improves navigation success rates by up to 30\%, substantially reduces trajectory errors against strong baselines, generalizes zero-shot to the unseen TartanDrive dataset, and scales naturally to high-dimensional humanoid navigation. These results position UniWM as a principled step toward unified, imagination-driven embodied navigation. The code and models are available at \url{https://github.com/UWMILab/UniWM}.
  
}
  
\date{June 30, 2026}

\metadata[Code]{\url{https://github.com/UWMILab/UniWM}}

\begin{document}

\maketitle

\input{sections/1_intro}
\input{sections/2_related_work}
\input{sections/3_method}
\input{sections/4_experiments}
\input{sections/5_conclusion}

\section*{Acknowledgments}
This work was supported in part by the University of Washington Faculty Startup Fund, the Carwein--Andrews Ph.D. Fellowship, the UW Graduate School Top Scholar Award, and PacTrans seed and small project funding through the U.S. Department of Transportation Region 10 University Transportation Center.

\appendix
\input{sections/X_appendix}

\clearpage\clearpage
\bibliographystyle{assets/plainnat}
\bibliography{main}

\end{document}

%% file: sections/1_intro.tex
\section{Introduction}
\label{sec:intro}

Visual navigation is a core capability for embodied AI and autonomous systems \citep{mirowski2016learning, chaplot2020learning, fu2022coupling, sridhar2024nomad}, enabling agents to interpret egocentric observations and sequentially choose actions to reach goals in complex environments \citep{karnan2022socially,yue2025video,hu2023gaia-1}.  It supports real-world applications such as robotic delivery, autonomous driving, and assistive technologies \citep{survey_3d_4d_world_models,worldlens,survey_vla4ad,xu2025U4D,li2026toward,liu2026driveva,liu2026palm}, where robust perception, accurate planning, and the ability to \emph{anticipate} how the environment will evolve under candidate actions are essential. Humans excel at this by mentally simulating future outcomes to plan in both familiar and novel settings~\citep{bar2025navigation,bartoccioni2025vavim_vavam}.

\begin{figure}[!t]
\setlength{\abovecaptionskip}{4pt}
\setlength{\belowcaptionskip}{2pt}
    \centering
    \includegraphics[width=\linewidth]{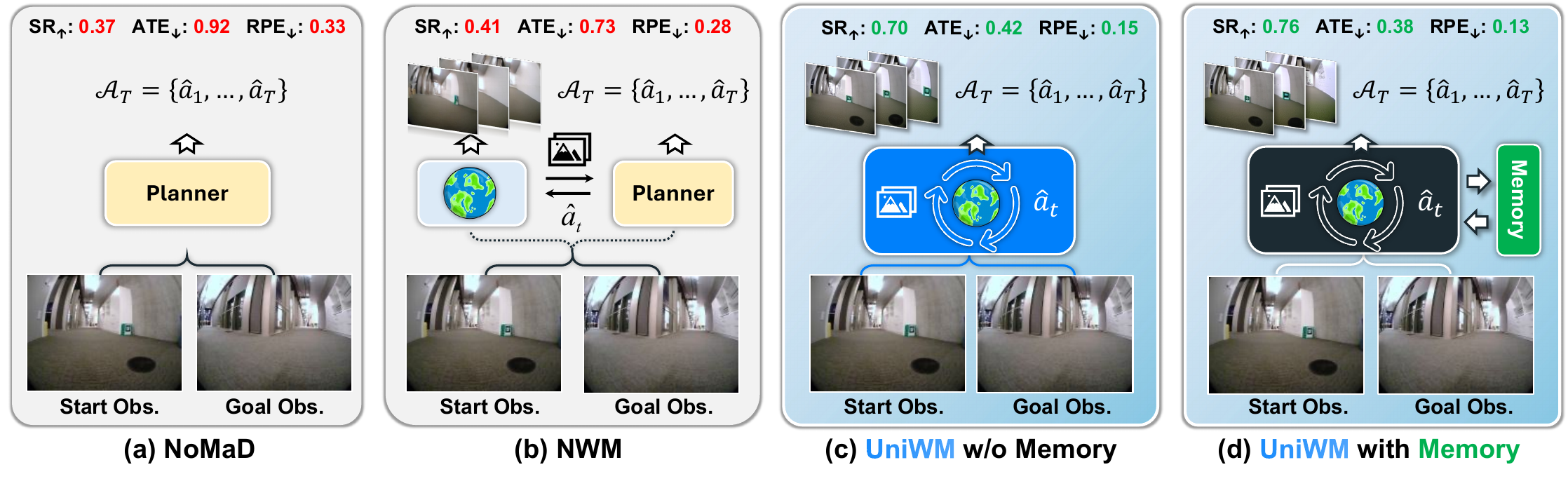}
    \caption{\small 
    \textbf{Comparisons among goal-conditioned visual navigation methods.}
    \textbf{(a)}~Navigation policy methods like NoMaD~\citep{sridhar2024nomad} directly predict action sequences $A_T$.
    \textbf{(b)}~World model for navigation like NWM~\citep{bar2025navigation} uses a world model to visualize future observations, enhancing a separate navigation planner.
    \textbf{(c)}~\ours~(no memory) unifies planning and visualization within one multimodal backbone, and actions are grounded in the imagined next observation while generating $A_T$ autoregressively.
    \textbf{(d)}~\ours~(with hierarchical memory) adds intra-step and cross-step memory banks, stabilizing longer-horizon rollouts and consistently yielding the highest SR and lowest errors. All panels use the same start/goal observations; headers report navigation SR$\uparrow$, ATE$\downarrow$, and RPE$\downarrow$ on HuRoN~\citep{hirose2023sacson}.}
    \label{fig:highlight}
\end{figure}

Despite rapid progress, current visual navigation systems are limited in fundamental ways (Fig.~\ref{fig:highlight}). \textbf{(a) Direct policy methods} (\eg, GNM~\citep{shah2022gnm}, VINT~\citep{shah2023vint}, NoMaD~\citep{sridhar2024nomad}) map observations to actions, but are tightly coupled to training distributions and often struggle to adapt in novel environments~\citep{song2025survey}. \textbf{(b) Modular pipelines} pair a planner with a separate world model: NavCoT~\citep{lin2024navcot} textualizes future observations and loses spatial fidelity, while NWM~\citep{bar2025navigation} uses diffusion-based rollouts and ranking. However, when prediction and control are learned in isolation and trajectory memory is absent, state-action misalignment arises and errors compound under partial observability and long horizons~\citep{ding2024understanding,xiao2025worldmem}. \textbf{(c) Unified autoregressive frameworks} provide a more principled direction by interleaving \emph{``imagining the next view''} with \emph{``predicting the next action''}, grounding decisions in anticipated outcomes and reducing misalignment (Fig.~\ref{fig:highlight}c). Yet, unification alone does not prevent gradual drift in longer-horizon reasoning. \textbf{(d) Hierarchical memory} supplies the missing inductive bias: retaining both immediate perceptual cues and longer-range trajectory context promotes temporal coherence, yielding higher SR and lower errors in challenging settings (Fig.~\ref{fig:highlight}d).

In short, effective navigation requires not only the ability to \emph{imagine while acting} but also to \emph{remember over time}. The central challenge is therefore to unify planning and imagination within a single backbone while injecting temporal structure to sustain stable long-horizon performance.

To address this challenge, we propose \ours, a unified memory-augmented world model that integrates navigation planning and visual imagination within a single multimodal autoregressive backbone (Fig.~\ref{fig:overall1}; Sec.~\ref{sec:foundations-formulation}). During training, we interleave planner and world-model samples and jointly optimize bin-token classification for actions and reconstruction for images in a shared tokenization space spanning actions, text, pose, and vision; the framework scales naturally with parameter-efficient tuning such as LoRA (Fig.~\ref{fig:overall1}a; Sec.~\ref{sec:unified-training}). At inference, UniWM alternates between predicting the next action and imagining the next egocentric view, explicitly grounding control in predicted visual outcomes and mitigating state--action misalignment (Fig.~\ref{fig:overall1}b; Sec.~\ref{sec:inference-memory}). We further introduce a two-level hierarchical memory that combines an intra-step cache with a cross-step trajectory store, augmenting attention via similarity gating and temporal decay to maintain coherent long-horizon rollouts and improve stability (Fig.~\ref{fig:detail}). Together, UniWM unifies planning and imagination within one backbone and offers a practical recipe for memory-augmented foresight in visual navigation.

Empirically, UniWM improves Success Rate and reduces ATE/RPE on Go Stanford~\citep{hirose2018gonet}, ReCon~\citep{shah2021rapid}, SCAND~\citep{karnan2022socially}, and HuRoN~\citep{hirose2023sacson} versus GNM~\citep{shah2022gnm}, VINT~\citep{shah2023vint}, NoMaD~\citep{sridhar2024nomad}, Anole-7B~\citep{chern2024anole}, and NWM~\citep{bar2025navigation}. For instance, Go Stanford SR rises from 0.45 to 0.75 (Table~\ref{tab:nav_metrics}; Fig.~\ref{fig:visual_result}). UniWM also generalizes zero-shot to unseen TartanDrive (SR 0.42; Table~\ref{tab:nav_metrics_unseen}; Fig.~\ref{fig:tartan}) and scales to 25-DoF humanoid navigation on 1X Humanoid (Table~\ref{tab:nav_metrics_humanoid}; Fig.~\ref{fig:humanoid}). Beyond navigation, UniWM improves one-step and rollout visualization (higher SSIM/PSNR, lower LPIPS/DreamSim; Table~\ref{tab:method_comparison}). Ablations attribute gains to reconstruction for imagination fidelity (and downstream navigation), bin-token loss for action accuracy, and hierarchical memory for long-horizon stability, with additional effects from token budget, joint training, memory-layer choice, goal conditioning, and substep interleaving (Tables~\ref{tab:context_ablation_combined}--\ref{tab:step_nav_metrics}; Fig.~\ref{fig:nav_vis_dual}).

In summary, this work provides the following key contributions:
\begin{itemize}
    \item \textbf{Unified architecture.} We propose \ours, to our knowledge, the \textbf{first} unified, memory-augmented world model that integrates visual navigation planning and imagination within a single multimodal autoregressive backbone, addressing the representational fragmentation of modular pipelines.

    \item \textbf{Unified training.} We introduce an end-to-end interleaved training strategy that unifies planner and world-model instances within one autoregressive backbone, jointly optimizing discretized action prediction and visual reconstruction to tightly align imagination with control.

    \item \textbf{Hierarchical memory.} We develop a hierarchical memory mechanism that fuses short-term perceptual cues with longer-term trajectory context via similarity-based retrieval and temporal weighting, enabling stable and coherent predictions over extended horizons.

    \item \textbf{Comprehensive validation.} Extensive experiments demonstrate consistent gains across benchmarks, stronger imagination fidelity, robust generalization to novel cases, and scalability to high-dimensional humanoid navigation.
\end{itemize}

%% file: sections/2_related_work.tex
\section{Related Work}
\label{sec:related-work}
World models have emerged as a unifying paradigm for learning predictive representations of environment dynamics, supporting simulation, decision-making, \etc. We summarize \textbf{two lines of research}: \textbf{(a)} advances in generic world-model architectures, and \textbf{(b)} world models for goal-conditioned visual navigation.

\noindent \textbf{World Modeling} has evolved from compact recurrent dynamics~\citep{ha2018world,hafner2019dream,hafner2022masteringataridiscreteworld,hafner2024masteringdiversedomainsworld} through Transformer-based designs (\eg, I-JEPA~\citep{assran2023self}, V-JEPA~\citep{bardes2024revisiting}, DINO-WM~\citep{baldassarre2025back}) to large-scale generative systems~\citep{google2024genie2,google2025genie3,xiong2026actworld,chu2026agentic,yu2026elsa3d}. Diffusion generators (Sora~\citep{brooks2024video}, Cosmos~\citep{agarwal2025cosmos}, Genie~\citep{bruce2024genie}) enable high-fidelity simulation and planning~\citep{alonso2024diffusion,valevski2024diffusion,bar2025navigation,mei2025vision,zhang2025epona,bian2025dynamiccity,robosense_challenge_2025,zhu2025spiral,xiong2026unidrive,chi2026driver,liu2026universe}, but often at the cost of efficiency and limited policy integration~\citep{xiao2025worldmem}. LLM-based approaches simulate dynamics via prompting~\citep{zhao2025drivedreamer,xing2025critiquesworldmodels} or improve spatial reasoning of VLMs through preference optimization~\citep{shen2026fine}, yet suffer from modality misalignment and memory degradation over long horizons. MVoT~\citep{li2025imagine} generates intermediate image visualizations for spatial reasoning~\citep{yu2025core3d} but is limited to narrow 2D scenarios without memory. WorldVLA~\citep{cen2025worldvla} unifies action and world modeling for robotic manipulation but predicts action sequences in a single step, forgoing intermediate imagination and trajectory-level memory.

In contrast, UniWM introduces structured memory into a unified multimodal backbone, enabling agents to both imagine while acting and remember over time, which jointly addresses alignment and stability issues of modular designs.

\noindent \textbf{Goal-Conditioned Navigation} is a natural testbed for world models, as it requires tight coupling between perception and policy~\citep{frey2023fast,meta2025embodied,wu2025govig}. Policy-centric methods~\citep{shah2022gnm,shah2023vint,sridhar2024nomad} map observations directly to actions without modeling environment dynamics. Navigation-oriented world models instead predict future observations to support temporally informed planning~\citep{yao2025navmorph,dong2026language}: PathDreamer~\citep{koh2021pathdreamer} used GAN-based simulation but relied on auxiliary inputs (\eg, semantic maps), limiting generalization~\citep{lin2024navcot}; NWM~\citep{bar2025navigation} integrates video prediction into the navigation loop and selects actions via CEM-based trajectory optimization; while CEM scores complete candidate trajectories, it still proposes actions independently of the world model's visual dynamics, leaving planning and perception loosely coupled. In response, we propose a unified multimodal backbone that aligns action prediction with observation imagination, enabling end-to-end navigation through temporally grounded dynamics modeling.

%% file: sections/3_method.tex
\section{Methodology}
\label{sec:method}
We present \ours, a unified, memory-augmented world model that performs \emph{planning} and \emph{visualization} within a single autoregressive multimodal backbone.

We \textbf{first} introduce preliminaries that replace the disjoint planner-world-model pair with one multimodal LLM augmented by hierarchical memory (Sec.~\ref{sec:foundations-formulation}). We \textbf{then} detail unified training, including multimodal tokenization and role-specific objectives for planning and world modeling (Sec.~\ref{sec:unified-training}), and \textbf{finally} describe hierarchical memory for stable long-horizon rollouts at inference time (Sec.~\ref{sec:inference-memory}).
\vspace{-0.1cm}

\begin{figure}[!t]
    \centering
    \includegraphics[width=\linewidth]{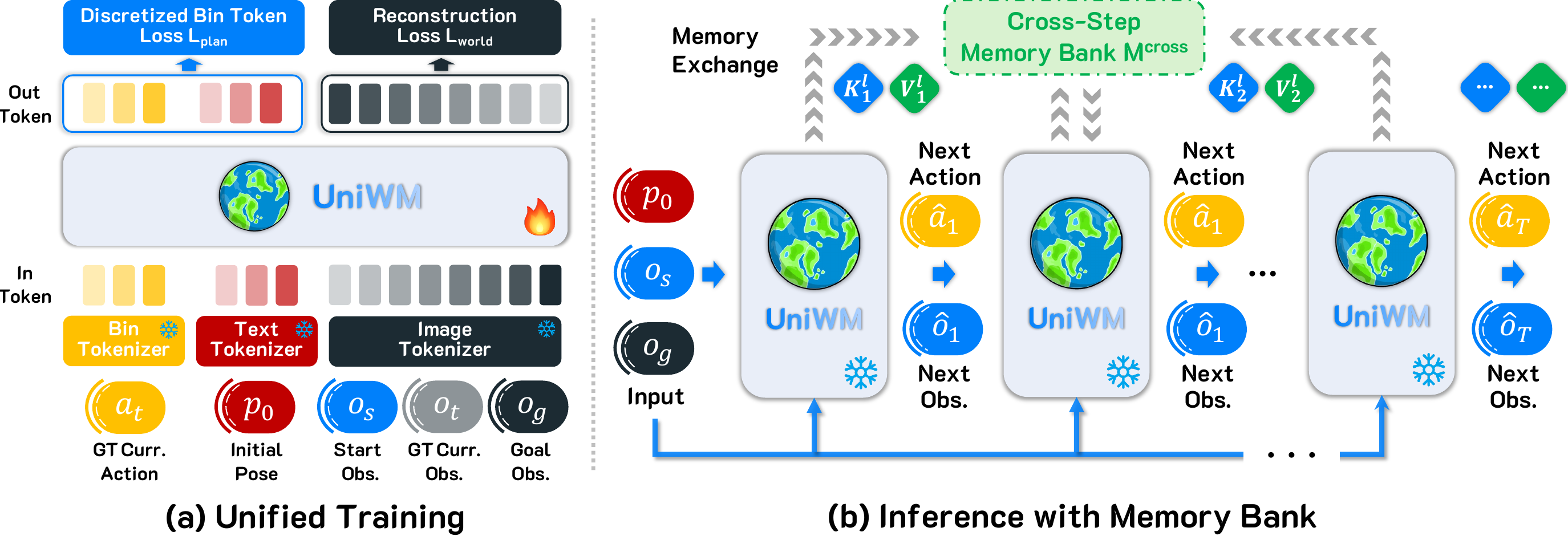}
    \caption{\small \textbf{Overview of the \ours~framework.}
    \textbf{(a) Training:} planner and world-model samples are interleaved within a single unified multimodal autoregressive backbone, optimized jointly with the discretized bin-token loss $\mathcal{L}_{\text{plan}}$ and the reconstruction loss $\mathcal{L}_{\text{world}}$; bin/text/image tokenizers map actions, pose, and observations to tokens.
    \textbf{(b) Inference:} a hierarchical memory supplies intra- and cross-step KV states ($\mathcal{M}^{\text{intra}}_{t}$ caches the current observation; $\mathcal{M}^{\text{cross}}_{t}$ accumulates prior steps) to augment attention, yielding robust trajectory-consistent alternating predictions of $\hat{a}_t$ (next action) and $\hat{o}_t$ (next observation). See Fig.~\ref{fig:detail} for the detailed memory mechanism.}
    \label{fig:overall1}
\end{figure}

\subsection{Preliminaries \& Unified Formulation}
\label{sec:foundations-formulation}
Given an egocentric RGB observation $o_s$ at the start, the initial agent pose $p_0 \in \mathbb{R}^3$ (position and yaw), and a goal observation $o_g$, the agent predicts a sequence of navigation actions $A_T=\{\hat{a}_1,\hat{a}_2,\dots,\hat{a}_T\}$ to reach the goal~\citep{sridhar2024nomad}. Each action $\hat{a}_t$ is either a continuous control command $(\mathbf{u}_t,\phi_t)$ or a terminal \texttt{Stop}, where $\mathbf{u}_t\in\mathbb{R}^2$ denotes planar translation (forward/backward, left/right) and $\phi_t\in\mathbb{R}$ denotes yaw rotation~\citep{bar2025navigation}. Actions are executed sequentially, and the agent must make monotonic progress toward $o_g$ until issuing \texttt{Stop}.

\textbf{World Models for Navigation.}~World models~\citep{ha2018world} predict future environment states (often image frames or video segments) conditioned on the current state and context:
$\hat{s}_{t+1} = \mathcal{W}(\hat{s}_t, \mathbf{c})$, where $\hat{s}_t$ is the current state, $\hat{s}_{t+1}$ is the predicted next state, and $\mathcal{W}$ is the learned dynamics model. The context $\mathbf{c}$ may include the executed action $a_t$, natural-language instructions, observation history, or other factors~\citep{russell2025gaia}. In navigation, world models $\mathcal{W}$ serve as imagination engines that anticipate future observations to guide planning.

A common instantiation couples two modules~\citep{bar2025navigation}: a \textbf{planner} that selects the next action given the current observation and the goal, and a \textbf{world model} that simulates the consequent observation conditioned on the chosen action and global cues such as the start and goal views:
\begin{equation}
    \setlength\abovedisplayskip{2pt}
    \setlength\belowdisplayskip{2pt}
\hat{a}_{t+1} = \mathcal{P}(\hat{o}_t, o_s, o_g), \quad
\hat{o}_{t+1} = \mathcal{W}(\hat{o}_t, \hat{a}_{t+1}, o_s, o_g),
\end{equation}
where $\hat{o}_t$ is the current observation, $\hat{a}_{t+1}$ is the action proposed by the planner $\mathcal{P}$, and $\hat{o}_{t+1}$ is the next observation visualized by $\mathcal{W}$. The start and goal observations $(o_s, o_g)$ provide the global navigation context. The two modules operate in a closed loop: $\mathcal{P}$ selects $\hat{a}_{t+1}$ conditioned on $\hat{o}_t$ and $(o_s, o_g)$, while $\mathcal{W}$ predicts $\hat{o}_{t+1}$ given $\hat{o}_t$ and $\hat{a}_{t+1}$, which is then fed back into $\mathcal{P}$. This iterative cycle enables imagination-based planning, allowing agents to simulate prospective action-observation trajectories before execution in the real environment. However, the modular training of $\mathcal{P}$ and $\mathcal{W}$ often leads to state–action misalignment, i.e., a discrepancy between the planner's implicit state assumption and the world model's predicted state, which degrades performance in complex and partially observable settings.

\textbf{Unified World Model with Memory.} To address these limitations, we replace the modular pair $(\mathcal{P},\mathcal{W})$ with a single multimodal backbone, UniWM, that tightly couples planning and visualization. UniWM is augmented with a hierarchical memory bank $\mathcal{M}_t=\{\mathcal{M}^{\mathrm{intra}}_t,\mathcal{M}^{\mathrm{cross}}_t\}$ that fuses short-term evidence with longer-range trajectory context (Fig.~\ref{fig:overall1}). At each step, UniWM performs:
\begin{equation} 
    \setlength\abovedisplayskip{2pt}
    \setlength\belowdisplayskip{2pt}
(\hat{a}_{t+1},\, \hat{o}_{t+1}) 
\;=\; 
\ours~\!\big(\hat{o}_t,\, o_s,\, o_g,\, p_0,\, \mathcal{M}_t\big), 
\label{eq:unified-step} 
\end{equation} 
where UniWM alternates between two substeps within the same MLLM backbone $F_{\theta}$: (i) \emph{action prediction} and (ii) \emph{navigation imagination}. Both are executed by $F_{\theta}$ and jointly learned via interleaved planner and world-model training with tailored objectives (Fig.~\ref{fig:overall1}a; Sec.~\ref{sec:unified-training}). During inference, hierarchical memory augments attention by integrating immediate evidence with longer-horizon context (Fig.~\ref{fig:overall1}b; Sec.~\ref{sec:inference-memory}), improving temporal coherence across rollouts.
\begin{enumerate}
[label=$\bullet$,leftmargin=1.1em,itemsep=2pt,topsep=2pt]
    \item \textbf{Navigation Planner (Action Prediction):} Given current observation $\hat{o}_t$, conditioned on start and goal observations ($o_s$, $o_g$), initial pose $p_0$, and memory bank $\mathcal{M}_t$, $F_{\theta}$ predicts next action $\hat{a}_{t+1}$: 
    \begin{equation}
    \setlength\abovedisplayskip{2pt}
    \setlength\belowdisplayskip{2pt}
      \hat{a}_{t+1} = F_{\theta}(\hat{o}_t, o_s, o_g, p_0, \mathcal{M}_t).
    \label{eq: nav-planner}
    \end{equation}
    \item \textbf{World Model (Navigation Visualization):}~Given current observation $\hat{o}_t$ and action $\hat{a}_{t+1}$, conditioned on $(o_s, o_g)$, $p_0$, and $\mathcal{M}_t$, $F_{\theta}$ predicts the next observation $\hat{o}_{t+1}$ after executing $\hat{a}_{t+1}$:
    \begin{equation}
        \setlength\abovedisplayskip{2pt}
    \setlength\belowdisplayskip{2pt}
    \hat{o}_{t+1} = F_{\theta}(\hat{o}_t, \hat{a}_{t+1}, o_s, o_g, p_0, \mathcal{M}_t).
    \label{eq: world-model}
    \end{equation}
\end{enumerate}
This design allows $F_{\theta}$ to act jointly as a \textbf{navigation planner} and a \textbf{world model}, alternating between roles until a terminal \texttt{Stop} is issued. During training, planner and world-model samples are interleaved so that $F_{\theta}$ learns both behaviors within a single autoregressive framework. At inference, a hierarchical memory bank augments $F_{\theta}$ by caching key--value states at both intra- and cross-step levels, enabling the integration of immediate observations with longer-range trajectory context. This unified formulation ensures consistent, memory-augmented world modeling throughout navigation.

\subsection{Unified Training Scheme}
\label{sec:unified-training}
Next, we describe how \ours~is trained as an autoregressive MLLM over text and image tokens. We build on Chameleon and Anole~\citep{team2024chameleon, chern2024anole}, which use a unified causal Transformer to jointly model multimodal token sequences (Fig.~\ref{fig:overall1}a).

\textbf{Data Preprocessing.} Each navigation trajectory provides two complementary sample types aligned with Eq.~\ref{eq: nav-planner} and Eq.~\ref{eq: world-model}. For the \textbf{navigation planner}, a sample contains $(o_s,o_g,o_t,p_0)$ with target $\hat{a}_{t+1}$. For the \textbf{world model}, the input additionally includes $a_{t+1}$ and the target becomes $\hat{o}_{t+1}$. Visual observations are inserted as \texttt{<image>} placeholders in structured multimodal prompts, and we use a sliding window to extract multiple samples per trajectory. Refer to \textbf{Appendix} for prompt design and examples. During training, planner and world-model samples are interleaved within the same batch to promote shared representations across roles.

\textbf{Multimodal Tokenization.}~We employ three tokenizers to unify visual and textual inputs. Following~\citep{gafni2022make, team2024chameleon}, a vector-quantized (VQ) image tokenizer discretizes images ($o_s$, $o_g$, $o_t$) into visual tokens via a learned codebook, while a byte-pair encoding (BPE) tokenizer~\citep{team2024chameleon} encodes pose $p_0$ and text prompts into text tokens. Actions $a_t$ are mapped to discrete bin tokens using the bin tokenizer, which we discuss below. The resulting token sequences are fed to a causal Transformer for joint multimodal modeling.

\textbf{Training Objective.}~To optimize our model for the distinct characteristics of the navigation planner and world model, we introduce tailored training objectives. At each iteration, our autoregressive MLLM jointly processes samples from both roles, producing logits across the unified vocabulary.

\textbf{$\bullet$ Discretized Bin Token Loss (Navigation Planner).}~We propose a new classification-based approach for training the planner, which formulates continuous action prediction as multi-class classification over discretized motion bins. Each navigation action $a_t \in \mathbb{R}^3$ is represented as $(x_t, y_t, \phi_t)$, where $x_t$ and $y_t$ denote planar translations and $\phi_t$ denotes yaw rotation. We uniformly partition each dimension into fixed-size bins with size $b = 0.01$, computing bin index as $\lfloor |v| / b \rfloor$ for value $v$. We use separate positive and negative token prefixes to encode the sign and another prefix for the target dimension. For example, $x$-axis translation with $v = 0.03$ is encoded as \texttt{<dx\_pos\_bin\_03>}. This scheme represents all three dimensions as special bin tokens from disjoint token sets: $\mathcal{T}_x$, $\mathcal{T}_y$, and $\mathcal{T}_\phi$. Let $P(t_i)$ denote the model's predicted distribution over all vocabulary tokens at decoding position $i$. We supervise the planner using \textbf{discretized bin token loss} over each action dimension:
\begin{equation}
\setlength\abovedisplayskip{3pt}
\setlength\belowdisplayskip{3pt}
\mathcal{L}_{\text{plan}}
= \frac{1}{3} \sum\nolimits_{k \in \{x,y,\phi\}}
\Big( - \log P\big(t_i = t_k^* \,\big|\, t_i \in \mathcal{T}_k \big) \Big) + \mathcal{L}_\text{CE},
\end{equation}
where $t^*_k$ is the ground-truth bin token in dimension $k$, and $\mathcal{L}_\text{CE}$ is the cross-entropy loss for output text tokens as output may also include text action \texttt{Stop}.

\textbf{$\bullet$ Reconstruction Loss (World Model).}~We introduce a \textbf{reconstruction loss} to enforce fidelity in the predicted future observations to encourage accurate navigation visualization. Given ground-truth visual embedding $\mathbf{v}_i$ for token $i$ (out of $n$ tokens in the next observation $\hat{o}_{t+1}$) and the visual codebook embeddings $\mathcal{E}=\{\mathbf{v}_1, \dots, \mathbf{v}_N\}$ where $N$ is the total number of visual token vocabulary:
\begin{equation}
\setlength\abovedisplayskip{2pt}
\setlength\belowdisplayskip{2pt}
\mathcal{L}_{\text{world}} = \frac{1}{n}\sum\nolimits_{i=1}^{n} \|\mathbf{v}_i, \mathcal{E}\|^2 \cdot P(t_i),
\end{equation}
where $\sum\nolimits_{i=1}^{n} \|\mathbf{v}_i, \mathcal{E}\|^2$ is the similarity vector indicating distances between $\mathbf{v}_i$ and all codebook embeddings, with lower similarity referring to larger distances, and $P(t_i)\in\mathbb{R}^{1\times N}$ denotes predicted probability distribution over visual tokens at position $i$. Throughout training, all tokenizers remain frozen, and only Transformer parameters are updated under autoregressive next-token prediction.

\begin{figure}[!t]
    \centering
    \includegraphics[width=\linewidth]{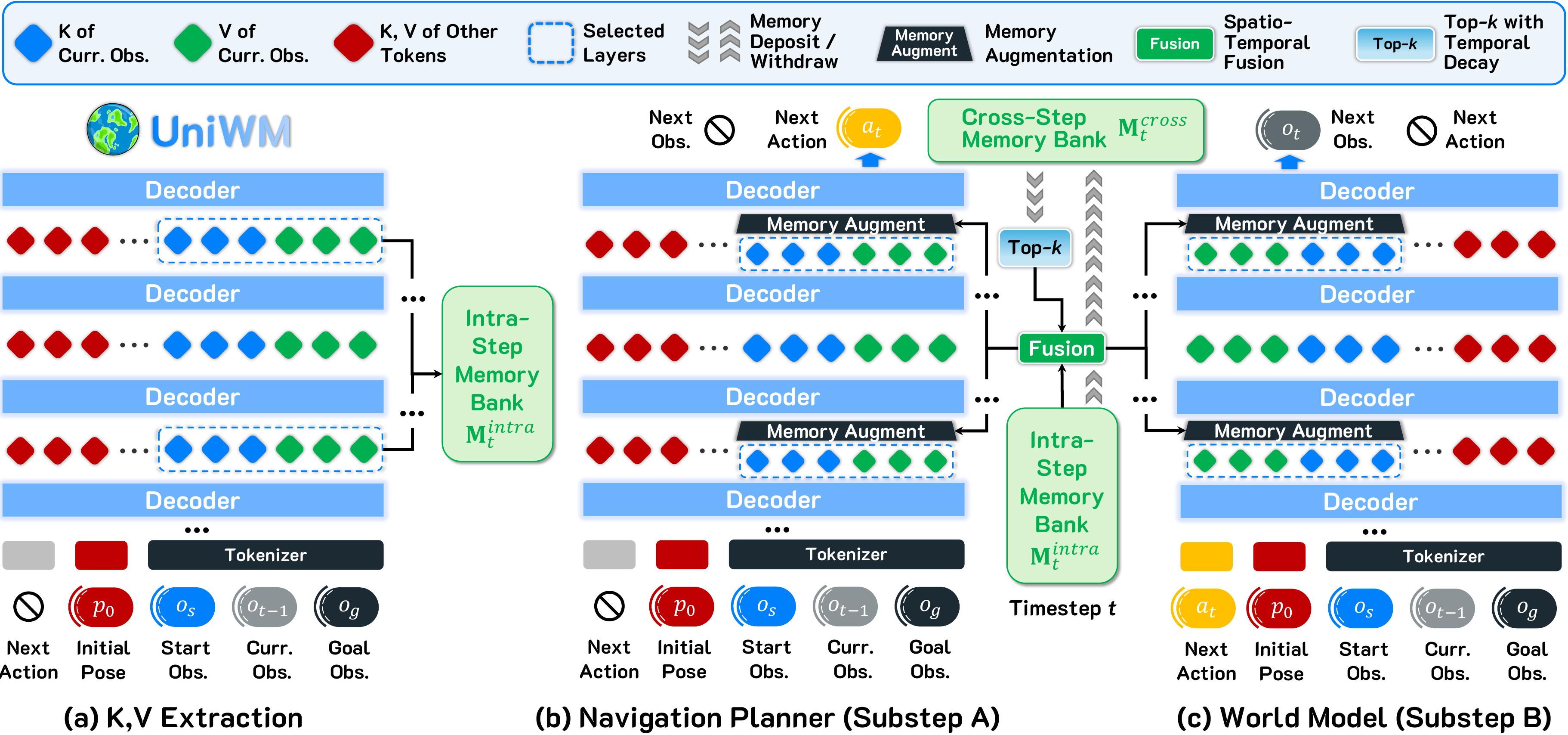}

    \caption{\small \textbf{Overview of hierarchical memory bank mechanism ($\mathcal{M}^{\text{intra}}_t$ \& $\mathcal{M}^{\text{cross}}_t$).}
    \textbf{(a)}~$KV$ (keys/values) extracted from selected layers are deposited into $\mathcal{M}^{\text{intra}}_t$ at the beginning of each step $t$ (Eq.~\ref{eq:seg_kv}). 
    \textbf{(b)(c)}~$\mathcal{M}^{\text{intra}}_t$ is merged with the accumulated cross-step memory $\mathcal{M}^{\text{cross}}_t$ via top-$k$ similarity gating (Eq.~\ref{eq:sim_gate_topk}) and exponential temporal decay (Eq.~\ref{eq:decay}), yielding a fused memory (Eq.~\ref{eq:fuse}) that augments attention for both the planner and the world-model substeps (Eq.~\ref{eq:enhance_atten}) to promote trajectory-consistent predictions. At the end of step $t$, $\mathcal{M}^{\text{intra}}_t$ (with timestamp $t$) is appended to $\mathcal{M}^{\text{cross}}_t$ for reliable reuse at step $t{+}1$, enabling robustly efficient rollouts.}
    \label{fig:detail}
\end{figure}

\subsection{Inference with Memory Bank}
\label{sec:inference-memory}
At the inference phase, \ours~alternates between two substeps: action prediction and navigation visualization. 
As illustrated in Fig.~\ref{fig:detail}, UniWM employs a hierarchical two-level memory bank mechanism. The \textbf{intra-step} memory $\mathcal{M}^{\text{intra}}_t$ caches key, value ($K$, $V$) pairs extracted from the current observation $\hat{o}_{t-1}$ at selected Transformer decoder layers, while the \textbf{cross-step} memory $\mathcal{M}^{\text{cross}}_t$ accumulates all past intra-step memories $\mathcal{M}^{\text{intra}}_{m}$, where $(m\in{1,...,t-1})$ together with their associated step indices $t_m$. 
This design allows $\mathcal{M}^{\text{cross}}_t$ to maintain a persistent trajectory-level context, thereby enabling $F_{\theta}$ to integrate both short-term and longer-term dependencies across steps.

\indent \textbf{Two-level Cache Design.}~At the beginning of each step $t$, the intra-step memory bank $\mathcal{M}^{\text{intra}}_t$ is reset to avoid contamination from the previous step: $\mathcal{M}^{\text{intra}}_t \leftarrow \varnothing$. Given the tokenized multimodal input, we identify the span of the current observation $\hat{o}_{t-1}$ by marking its token sequence with two special boundary tokens, $<$boss$>$ and $<$eoss$>$, thereby yielding the index set $\mathcal{I}_t$.  
We then extract $K, V$ pairs to form $\mathcal{M}^{\text{intra}}_t$ only from this span at a selected subset of decoder layers $L_{\text{save}} \subseteq L = \{l_0,\dots,l_{31}\}$:
\begin{equation}
        \setlength\abovedisplayskip{2pt}
    \setlength\belowdisplayskip{2pt}
\label{eq:seg_kv}
\mathcal{M}^{\text{intra}}_t = \{K^{(l)}_t,V^{(l)}_t\} \;=\; 
\big\{ f^{(l)}_{K}(\mathbf{x}_{\mathcal{I}_t}),~ f^{(l)}_{V}(\mathbf{x}_{\mathcal{I}_t}) \big\}, 
\quad where~~l \in L_{\text{save}},
\end{equation}

\noindent where $\{K^{(l)}_t,V^{(l)}_t\}$ denotes keys and values obtained from the $l$-th decoder layer at step $t$, 
$\mathbf{x}$ represents the hidden states of the multimodal input sequence at that layer, 
$\mathbf{x}_{\mathcal{I}_t}$ refers to the slice of hidden states indexed by $\mathcal{I}_t$, 
and $f^{(l)}_{K}$ and $f^{(l)}_{V}$ are the key and value projection mappings in layer $l$. In parallel, as demonstrated in Fig.~\ref{fig:detail}, the cross-step memory $\mathcal{M}^{\text{cross}}_t$ aggregates selected intra-step caches from previous $t{-}1$ steps with timestamps $t_m$: $\mathcal{M}^{\text{cross}}_t = \{(K^{(l)}_m,V^{(l)}_m,t_m)\}_{l \in L_{\text{save}}}$.  

\indent \textbf{Spatio-temporal Fusion.}~
At each action prediction substep of step $t$, the intra-step memory $\mathcal{M}^{\text{intra}}_t$ is merged with the accumulated cross-step memory $\mathcal{M}^{\text{cross}}_t$ to construct a fused memory $\tilde{\mathcal{M}}_t$, which subsequently enhances the attention mechanism for both substeps. 
This fusion incorporates spatial similarity selection and temporal recency weighting as shown in Fig.~\ref{fig:detail}:

\textit{(i) Similarity gating.}~We flatten both current and historical keys and compute entry-wise cosine similarity $s^{(l)}_m$. The indices of the top-$k$ most similar entries are collected into the set $h^{(l)}_t$:
\begin{equation}        \setlength\abovedisplayskip{2pt}
    \setlength\belowdisplayskip{2pt}
\label{eq:sim_gate_topk}
s^{(l)}_m \;=\; \operatorname{cos}\!\big(K^{(l)}_t,\,K^{(l)}_m\big), 
\quad h^{(l)}_t \;=\; \mathrm{top\text{-}}k\big(\,s^{(l)}_m\big), 
\quad \text{where}~~m \in \{1,\dots,t-1\}.
\end{equation}

\textit{(ii) Temporal decay.}  
Each selected entry is weighted by an exponential decay factor determined by its recency gap $\Delta t_m = t - t_m$, that larger weights correspond to a stronger influence on subsequent predictions. Here we set $\gamma=0.2$, which biases the weighting toward more recent steps:
\begin{equation}
\setlength\abovedisplayskip{1pt}
\setlength\belowdisplayskip{1pt}
\label{eq:decay}
\alpha^{(l)}_m \;=\; 
\frac{\exp\!\big(-\gamma\,\Delta t_m\big)}{\sum_{j\in h^{(l)}_t}\exp\!\big(-\gamma\,\Delta t_j\big)} \;.
\end{equation}

\textit{(iii) Memory fusion.}  
The fused memory $\tilde{\mathcal{M}}_t = \{\tilde{K}_{t}^{(l)}, \tilde{V}_{t}^{(l)}\}_{l\in L_{\text{save}}}$ is formed by concatenating the current intra-step memory with the weighted historical entries so that historical contributions are explicitly modulated by both spatial similarity and temporal recency. Thus, for $h \in h^{(l)}_t,~ l \in L_{\text{save}}$, we have:
\begin{equation}
        \setlength\abovedisplayskip{2pt}
    \setlength\belowdisplayskip{2pt}
\label{eq:fuse}
\tilde{K}^{(l)}_t=\textbf{Concat}~\!\big(K^{(l)}_t, \alpha^{(l)}_h K^{(l)}_h\big),~~
\tilde{V}^{(l)}_t=\textbf{Concat}~\!\big(V^{(l)}_t, \,\alpha^{(l)}_h V^{(l)}_h\big).
\end{equation}
\indent \textbf{Memory-Augmented Attention.}~
The fused memory $\tilde{\mathcal{M}}_t$ then directly engages in cross-attention computation. The attention mechanism can be formally described as scaled dot-product attention:
\begin{equation}
        \setlength\abovedisplayskip{2pt}
    \setlength\belowdisplayskip{2pt}
\label{eq:enhance_atten}
    \tilde{Q}^{(l)}_{t}= \text{Att}(Q^{(l)}_t, \tilde{K}^{(l)}_t, \tilde{V}^{(l)}_t) 
    = \text{softmax}\!\left(\frac{Q^{(l)}_t \tilde{K}^{(l)\top}_t}{\sqrt{d_k}}\right) \tilde{V}^{(l)}_t,
\end{equation}
where $Q^{(l)}_t$ denotes the current query at layer $l$, and $d_k$ is the key dimension. $\tilde{Q}^{(l)}_{t}$ subsequently propagate through later predictions. This mechanism equips UniWM with trajectory-consistent reasoning by leveraging both current observations and temporally structured historical memories.


\indent \textbf{Rollout Procedure.}~The full inference loop (detailed algorithm in \textbf{Appendix}) is: at each step $t$, UniWM resets $\mathcal{M}^{\text{intra}}_t$, extracts and fuses KV states via Eqs.~\ref{eq:seg_kv}--\ref{eq:fuse}, then alternates between action prediction (Eq.~\ref{eq: nav-planner}) and observation generation (Eq.~\ref{eq: world-model}) under memory-augmented attention (Eq.~\ref{eq:enhance_atten}). Current $\mathcal{M}^{\text{intra}}_t$ is then appended to $\mathcal{M}^{\text{cross}}_t$, and process repeats until a \texttt{Stop} action is emitted.

%% file: sections/4_experiments.tex
\section{Experiments}
\label{sec:exp}

\subsection{Experimental Settings}
\noindent \textbf{Datasets.}~We evaluate in two settings using six open-source datasets. 
\begin{itemize}
    \item \underline{\textbf{Setting~(i)}} \emph{Egocentric Navigation with Wheeled/Legged Robots}: \textbf{Go Stanford}~\citep{hirose2018gonet}, \textbf{ReCon}~\citep{shah2021rapid}, \textbf{SCAND}~\citep{karnan2022socially}, and \textbf{HuRoN}~\citep{hirose2023sacson} are used for training and in-domain evaluation, spanning indoor corridors to socially compliant outdoor paths. \textbf{TartanDrive}~\citep{triest2022tartandrive} is held out for unseen testing; its visible ego-robot structures induce a realistic distribution shift. 

    \item \underline{\textbf{Setting~(ii)}} \emph{Humanoid Navigation}: we use the navigation subset of the \textbf{1X Humanoid Dataset}~\citep{1X_Technologies_1X_World_Model_2024} with 25-DoF joint-angle actions. Following~\citep{bagchi2026walk}, we train a separate UniWM instance due to incompatible action representations. 
\end{itemize}
For Setting~(i), we normalize per-frame displacement by the average step size, filter backward motions~\citep{bar2025navigation, sridhar2024nomad} and trajectories shorter than three steps, and segment visual streams into sub-scenes via Qwen2.5-VL~\citep{bai2025qwen2}. For 1X Humanoid, we retain only navigation episodes. After preprocessing, the trajectory counts are: Go Stanford (4457/496), ReCon (4652/517), SCAND (2560/285), HuRoN (4642/516), 1X Humanoid (6599/733), and TartanDrive (eval only: 500).

\noindent \textbf{Evaluation Metrics.}
We report two suites of metrics. 
\textbf{(1) Navigation quality:} Absolute Trajectory Error (ATE), Relative Pose Error (RPE)~\citep{sturm2012evaluating}, and Success Rate (SR). SR counts success when the final distance to the goal is below the agent’s average step size (meters). 
\textbf{(2) Visualization quality:} SSIM~\citep{wang2004ssim}, PSNR~\citep{hore2010image}, LPIPS~\citep{zhang2018unreasonable}, and DreamSim~\citep{fu2023dreamsim}. To assess long-horizon stability under rollout, we also compute SSIM@n, PSNR@n, LPIPS@n, and DreamSim@n. Due to space limits, kindly refer to \textbf{Appendix} for additional details.

\noindent \textbf{Implementation Details.}~UniWM is fine-tuned from GAIR Anole-7B~\citep{chern2024anole} (4096-token context) with all tokenizers frozen. Images are resized to $448\times448$ and discretized into 784 visual tokens. We update only LoRA~\citep{hu2022lora} adapters (rank $=16$) on the Transformer's \emph{qkv} projections~\citep{liu2023llava}. Training uses AdamW for 20 epochs (lr $2\times10^{-4}$, batch size 8) on 4$\times$A100-80GB GPUs. At inference, two boundary tokens ($<$boss$>$/$<$eoss$>$, IDs 8196/8197) trigger KV deposit into the intra-step memory bank, and we extract KV states from decoder layers $\{l_0,l_7,l_{15},l_{23},l_{31}\}$. All baselines are \textbf{retrained from scratch} on the same four training datasets (Go Stanford, ReCon, SCAND, HuRoN), except Anole-7B (zero-shot prompting). All models are evaluated zero-shot on TartanDrive under identical conditions. Since Aether~\citep{zhu2025aether} does not release training code, we use its checkpoint and report it only for visualization, as it operates in a camera-pose action space incompatible with our robot-action benchmarks. Due to space limits, kindly refer to \textbf{Appendix} for additional implementation details.

\begin{table}[!t]
\setlength{\abovecaptionskip}{0pt}
\setlength{\belowcaptionskip}{0pt}
\centering
\caption{\small \textbf{Comparisons with SOTA Methods upon Goal-Conditioned Visual Navigation} on evaluation splits of Go Stanford, ReCon, SCAND, and HuRoN with SR, ATE, and RPE. Best results for each metric are in bold.}
\label{tab:nav_metrics}
\resizebox{\linewidth}{!}{%
\begin{tabular}{lcccccccccccc}
\toprule[1.2pt]
\multirow{2}{*}{\textbf{Method}} 
& \multicolumn{3}{c}{\textbf{Go Stanford}} 
& \multicolumn{3}{c}{\textbf{ReCon}} 
& \multicolumn{3}{c}{\textbf{SCAND}} 
& \multicolumn{3}{c}{\textbf{HuRoN}} \\
\cmidrule(lr){2-4} \cmidrule(lr){5-7} \cmidrule(lr){8-10} \cmidrule(lr){11-13}
& \textbf{SR $\uparrow$}& \textbf{ATE $\downarrow$}& \textbf{RPE $\downarrow$}& \textbf{SR $\uparrow$}& \textbf{ATE $\downarrow$}& \textbf{RPE $\downarrow$}& \textbf{SR $\uparrow$}& \textbf{ATE $\downarrow$}& \textbf{RPE $\downarrow$}& \textbf{SR $\uparrow$}& \textbf{ATE $\downarrow$}& \textbf{RPE $\downarrow$}\\
\midrule
GNM \citep{shah2022gnm}   
& 0.27     & 1.11     & 0.31     
& 0.72     &  0.70    & 0.20     
& 0.49     &  0.51    & 0.21     
& 0.36     & 1.07     &  0.35    \\
VINT \citep{shah2023vint}    
& 0.29     &  1.09    & 0.35     
& 0.68     & 0.84     &  0.28    
&  0.45    &  0.58    &  0.28    
&  0.30    & 1.19     & 0.43     \\
NoMaD \citep{sridhar2024nomad}   
&  0.33    & 0.94     &  0.30    
&  0.71    &  0.77    &  0.21    
&  0.50    &   0.54   & 0.23     
&  0.37    &   0.92   & 0.33      \\
Anole-7B \citep{chern2024anole}   
& 0.18     & 2.18     & 0.73     
& 0.41     & 1.74     & 0.69     
& 0.29     & 1.37     & 0.71     
& 0.20     & 1.92     & 0.78      \\
NWM \citep{bar2025navigation}   
&  0.45    & 0.80    & 0.27    
&  0.79    & 0.58    & 0.17    
&  0.55    & 0.41    & 0.19    
&  0.41    & 0.73    & 0.28      \\
\rowcolor{w_1!7}\textbf{\ours~(w/o $\mathcal{M}$)} 
& 0.71 & 0.32 & 0.10 
& 0.82 & 0.35 & 0.12 
& 0.61 & 0.36 & 0.14 
& 0.70 & 0.42 & 0.15 
\\
\rowcolor{w_1!15} \textbf{\ours~(w/ only $\mathcal{M}^{\text{intra}}_t$)} 
& 0.73 & 0.29 & \textbf{0.09} 
& 0.85 & 0.38 & 0.13 
& 0.64 & 0.33 & \textbf{0.13} 
& 0.74 & 0.44 & 0.15 
\\
\rowcolor{w_1!27}\textbf{\ours~(w/ $\mathcal{M}^{\text{intra}}_t$ \& $\mathcal{M}^{\text{cross}}_t$)} 
& \textbf{0.75} & \textbf{0.22} & \textbf{0.09} 
& \textbf{0.93} & \textbf{0.34} & \textbf{0.11} 
& \textbf{0.68} & \textbf{0.32} & \textbf{0.13} 
& \textbf{0.76} & \textbf{0.38} & \textbf{0.13} \\
\bottomrule[1.2pt]
\end{tabular}
}
\end{table}

\subsection{Comparison to State-of-the-Art Approaches}
\noindent \textbf{Navigation Performance.}~Table~\ref{tab:nav_metrics} reports goal-conditioned navigation results on four in-domain datasets (Go Stanford, ReCon, SCAND, HuRoN), and Fig.~\ref{fig:visual_result} shows qualitative comparisons (additional results are placed in the \textbf{Appendix}). We compare UniWM against policy-centric baselines GNM~\citep{shah2022gnm}, VINT~\citep{shah2023vint}, and NoMaD~\citep{sridhar2024nomad}, as well as Anole-7B~\citep{chern2024anole} under zero-shot prompting. We also include NWM~\citep{bar2025navigation}, which performs planning via MPC with a CDiT world model. UniWM consistently outperforms all baselines. Even without memory, UniWM yields large gains in SR and improves ATE/RPE across datasets. Adding intra-step memory further stabilizes predictions, while cross-step memory strengthens long-horizon consistency, producing the best overall performance.

\begin{figure}[t]
    \centering
    \includegraphics[width=\linewidth]{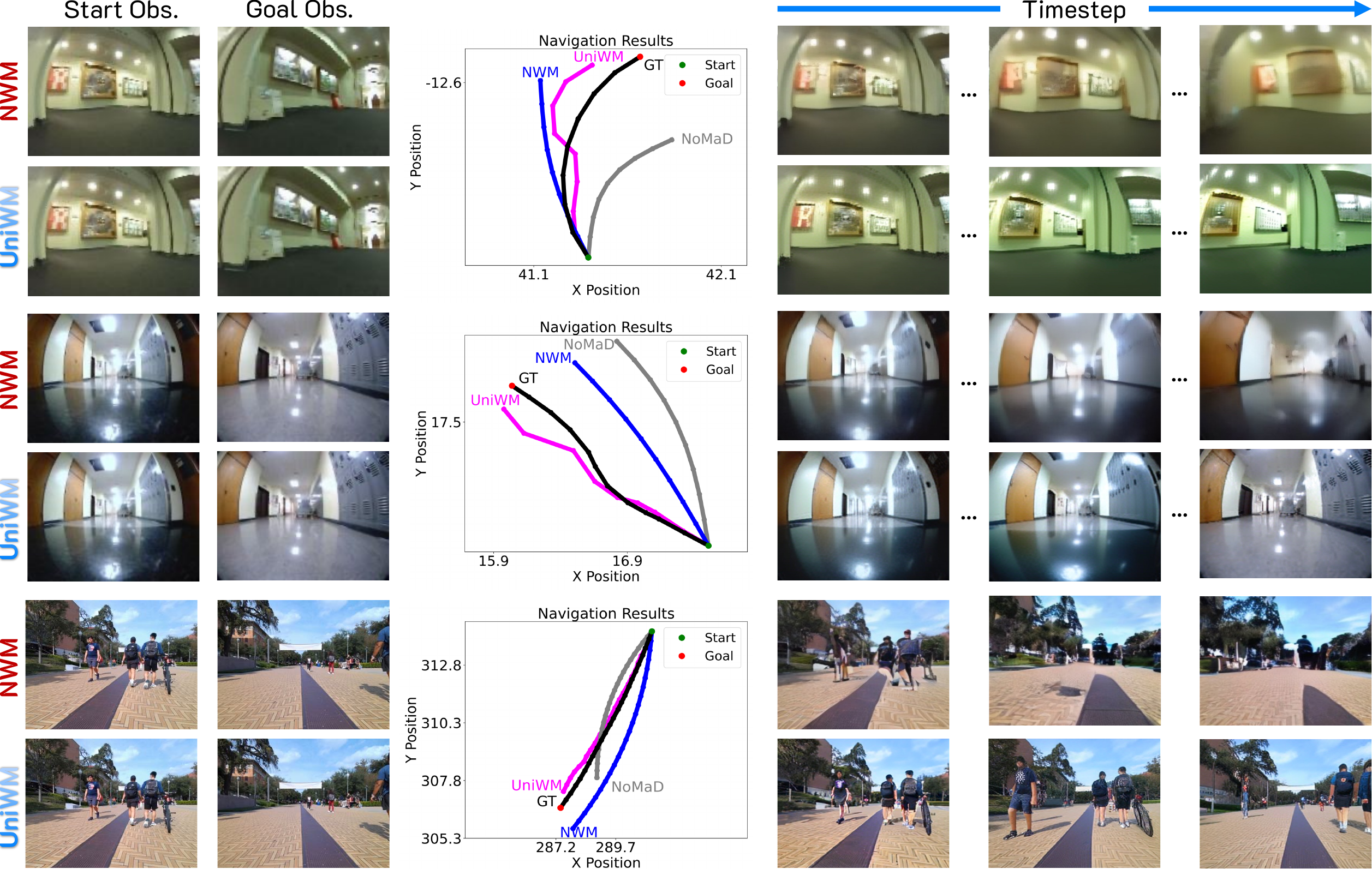}
    \vspace{-0.7cm}  
    \caption{\small \textbf{Qualitative Comparisons} on Go Stanford and HuRoN datasets across UniWM, NWM, and NoMaD. Qualitative results here include both static indoor environments and outdoor scenarios with moving pedestrians. The central trajectory plots highlight the difference between predicted $A_T$ and the ground-truth.}
    \label{fig:visual_result}
\end{figure}

\begin{table}[t]
\setlength{\abovecaptionskip}{0pt}
\setlength{\belowcaptionskip}{0pt}
\centering
\caption{\small \textbf{Comparisons with SOTA methods on visual quality assessment}, averaged over evaluation splits of Go Stanford, ReCon, SCAND, and HuRoN. Best results for each metric are in bold.}
\label{tab:method_comparison}
\resizebox{\linewidth}{!}{%
\begin{tabular}{lccccccccc}
\toprule
\textbf{Method} & \textbf{SSIM $\uparrow$} & \textbf{PSNR $\uparrow$} & \textbf{LPIPS $\downarrow$} & \textbf{DreamSIM $\downarrow$} & \textbf{SSIM$@5$ $\uparrow$} & \textbf{PSNR$@5$ $\uparrow$} & \textbf{LPIPS$@5$ $\downarrow$} & \textbf{DreamSIM$@5$ $\downarrow$} \\
\midrule
Diamond~\citep{alonso2024diffusion} 
& 0.311 & 9.837 & 0.410 & 0.131 
& 0.186 & 6.352  & 0.582 & 0.252 \\
Aether~\citep{zhu2025aether}
&0.373	&10.952  &0.271	&0.068	&0.235	&7.183  &0.514	&0.158 \\
NWM~\citep{bar2025navigation}     
&0.389  &11.420  &0.318  & 0.089 
&0.256  &7.755  & 0.494 & 0.174 \\
\rowcolor{w_1!27} \ours    
& \textbf{0.457} & \textbf{13.607} &\textbf{0.254}  & \textbf{0.041} 
&\textbf{0.350}  &\textbf{10.874}  &\textbf{0.435}  & \textbf{0.126} 
\\
\bottomrule
\end{tabular}}
\end{table}

\begin{table}[t]
\setlength{\abovecaptionskip}{0pt}
\setlength{\belowcaptionskip}{0pt}
\centering
\caption{\small \textbf{Impact of Context Size and Image Token Length} on both navigation and visualization performance, averaged over four datasets. All settings are evaluated without memory banks. Best results for each metric are highlighted in \textbf{bold}.}
\label{tab:context_ablation_combined}
\resizebox{1\linewidth}{!}{%
\begin{tabular}{ccccccccccc}
\toprule
\multirow{2}{*}{\textbf{Context}} & \multirow{2}{*}{\textbf{Token Len.}} 
& \multicolumn{3}{c}{\textbf{Navigation}} 
& \multicolumn{6}{c}{\textbf{Visualization}} \\
\cmidrule(lr){3-5} \cmidrule(lr){6-11}
& & \textbf{SR $\uparrow$} & \textbf{ATE $\downarrow$} & \textbf{RPE $\downarrow$}
& \textbf{SSIM $\uparrow$} & \textbf{LPIPS $\downarrow$} & \textbf{DreamSIM $\downarrow$} & \textbf{SSIM$@5$ $\uparrow$} & \textbf{LPIPS$@5$ $\downarrow$} & \textbf{DreamSIM$@5$ $\downarrow$} \\
\midrule
\rowcolor{w_1!27} 1 & 784 & \textbf{0.71}    & \textbf{0.36}    &  \textbf{0.13}   
& \textbf{0.457}  &0.254  & \textbf{0.041} 
&\textbf{0.350}   &\textbf{0.435}  & \textbf{0.126} \\
2 & 625 & 0.68    & 0.39    &  0.15   
& 0.448  & \textbf{0.247}  & 0.051  
& 0.336  & 0.451  & 0.137  \\
2 & 484 & 0.55    & 0.53    &  0.26   
& 0.365  & 0.328  & 0.084  
& 0.258  & 0.515  & 0.192  \\
4 & 484 & 0.64    & 0.44    &  0.19   
& 0.425  & 0.285  & 0.052  
& 0.315  & 0.462  & 0.141  \\
\bottomrule
\end{tabular}}
\end{table}

\begin{table}[t]
\setlength{\abovecaptionskip}{0pt}
\setlength{\belowcaptionskip}{0pt}
\centering
\caption{\small \textbf{Comparison of the unified UniWM and the separate planner + world-model setting} on both navigation (SR, ATE, RPE) and visualization metrics (SSIM, LPIPS, DreamSim). Best results for each metric are highlighted in \textbf{bold}.}
\label{tab:separate}
\resizebox{\linewidth}{!}{%
\begin{tabular}{lccccccccc}
\toprule
\textbf{Method} &\textbf{SR $\uparrow$}& \textbf{ATE $\downarrow$} & \textbf{RPE $\downarrow$} & \textbf{SSIM $\uparrow$} & \textbf{LPIPS $\downarrow$} & \textbf{DreamSIM $\downarrow$}  & \textbf{SSIM$@5$ $\uparrow$} & \textbf{LPIPS$@5$ $\downarrow$} & \textbf{DreamSIM$@5$ $\downarrow$} \\
\midrule
UniWM (Separate)  &0.65 &0.41 &0.16 &0.443 &0.280 &0.055 & 0.329     &0.470 & 0.154\\
\rowcolor{w_1!27}UniWM  & \textbf{0.71}  &  \textbf{0.36}   & \textbf{0.13}   &  \textbf{0.457} &  \textbf{0.254} &  \textbf{0.041} &  \textbf{0.350} &  \textbf{0.435}& \textbf{0.126}\\
\bottomrule
\end{tabular}}
\end{table}

\noindent \textbf{Visualization Performance.}~Table~\ref{tab:method_comparison} evaluates imagination quality. We compare with Diamond~\citep{alonso2024diffusion} (UNet), Aether~\citep{zhu2025aether} (geometry-aware model built on CogVideoX~\citep{yang2024cogvideox}), and NWM~\citep{bar2025navigation} (CDiT). UniWM is competitive across all metrics. For one-step prediction, it achieves the best SSIM and DreamSim. Under open-loop rollouts, UniWM remains stable with $\text{SSIM}@5=0.350$, preserving semantic consistency and reducing compounding errors over longer horizons.

\subsection{Ablation Study \& Analyses}

\noindent \textbf{1.~How do context size and token length affect performance?} 

Table~\ref{tab:context_ablation_combined} varies both factors under Anole-7B’s fixed 4096-token window, revealing a trade-off between temporal coverage and spatial resolution. Increasing either context or token length improves results (\eg, 2$\times$484$\rightarrow$4$\times$484 or 2$\times$484$\rightarrow$2$\times$625). Comparing 1$\times$784 with 2$\times$625 and 4$\times$484 suggests that, under a fixed token budget, spatial resolution contributes more than additional context frames.

\begin{figure}[!t]
\setlength{\abovecaptionskip}{-8pt}
\setlength{\belowcaptionskip}{0pt}
  \centering
  \begin{minipage}[t]{0.35\linewidth}
      \centering
      \includegraphics[width=\linewidth]{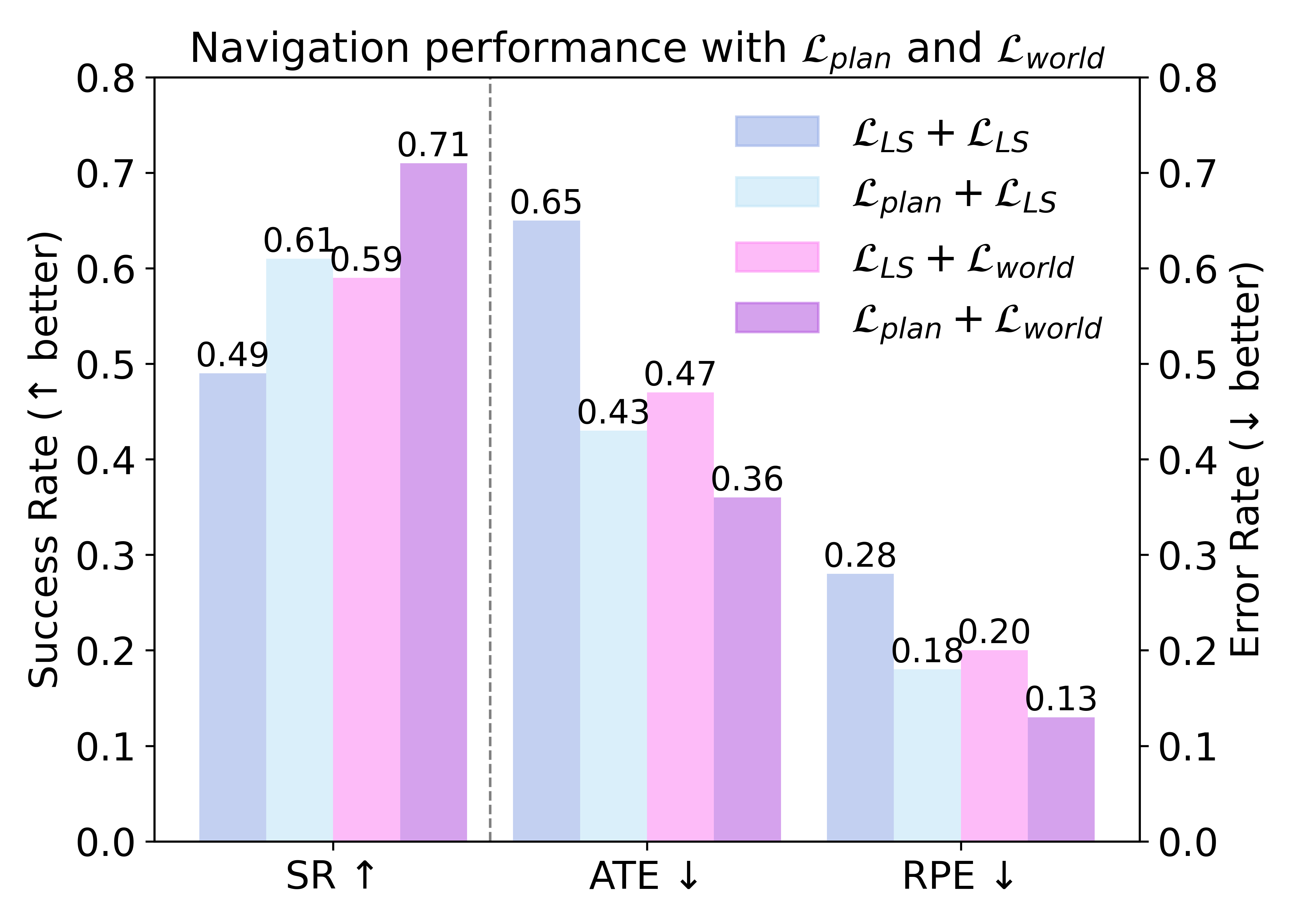}
      \label{fig:nav_ablation}
  \end{minipage}
  \hfill
    \begin{minipage}[t]{0.62\linewidth}
      \centering
      \includegraphics[width=\linewidth]{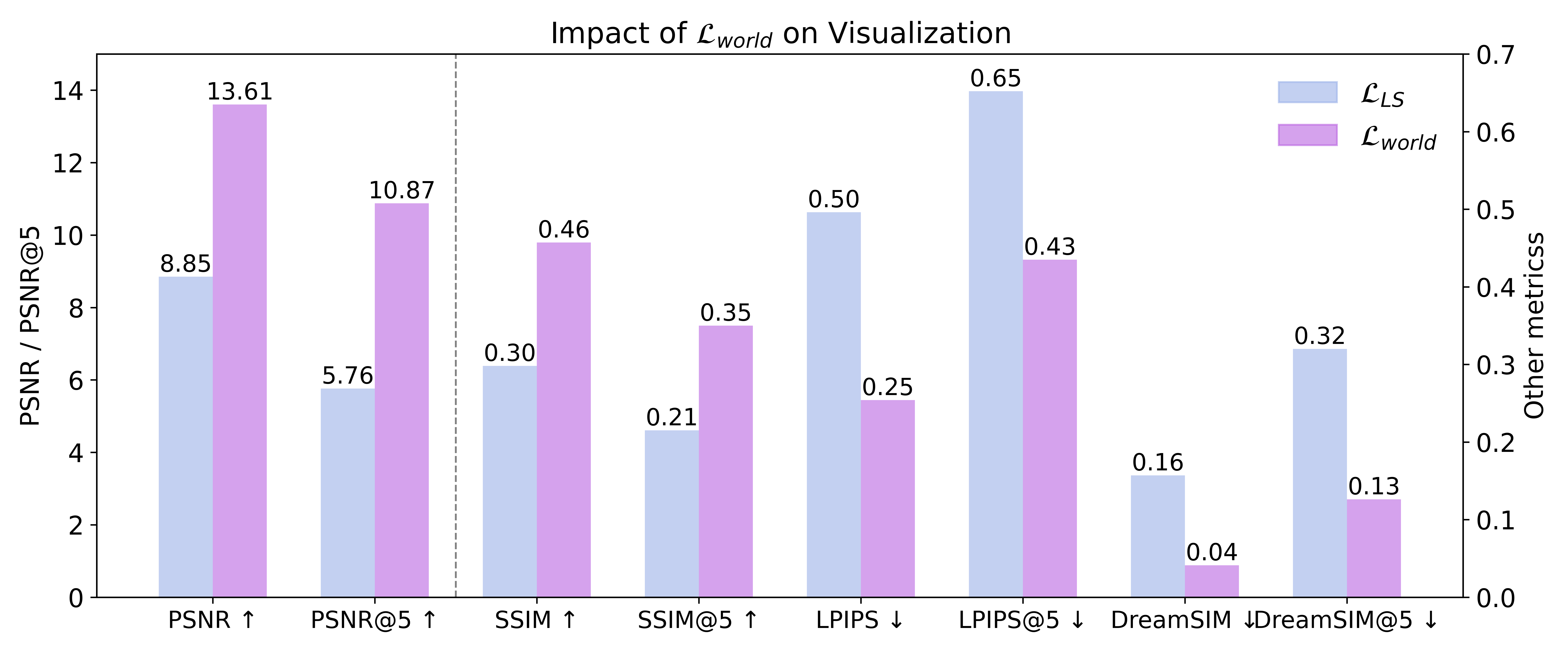}
      \label{fig:vis_ablation}
  \end{minipage}
  \caption{\small \textbf{Impact of discretized bin-token loss ($\mathcal{L}_{\text{plan}}$) and reconstruction Loss ($\mathcal{L}_{\text{world}}$)} on navigation (left) and visualization (right) performance, averaged over evaluation splits of Go Stanford, ReCon, SCAND, and HuRoN. X-axis arrows indicate whether higher or lower values are preferable.}
  \label{fig:nav_vis_dual}
\end{figure}

\noindent \textbf{2.~Do $\mathcal{L}_{\text{plan}}$ and $\mathcal{L}_{\text{world}}$ help training?} 

We evaluate both losses against a label-smoothing baseline $\mathcal{L}_{\text{LS}}$ (used only in this ablation). As shown in Fig.~\ref{fig:nav_vis_dual}, replacing $\mathcal{L}_{\text{LS}}$ with $\mathcal{L}_{\text{world}}$ substantially improves visualization quality (including rollouts) and also benefits navigation. Combining $\mathcal{L}_{\text{plan}}$ and $\mathcal{L}_{\text{world}}$ performs best overall: $\mathcal{L}_{\text{plan}}$ yields larger navigation gains than $\mathcal{L}_{\text{world}}$ (SR $+0.12$ vs.\ $+0.10$), suggesting that $\mathcal{L}_{\text{world}}$ helps navigation primarily via improved imagination, whereas $\mathcal{L}_{\text{plan}}$ directly optimizes action accuracy.

\noindent \textbf{3.~Should the navigation planner and world model be trained jointly?} 

Table~\ref{tab:separate} compares unified UniWM against a variant in which the planner and world model are trained separately with identical data and schedule. The unified model consistently outperforms across both navigation and visualization metrics, confirming that joint training more effectively aligns imagination with control.

\noindent \textbf{4.~Do we need both intra-step and cross-step memory at inference?} 

\noindent Table~\ref{tab:nav_metrics} compares three variants: no memory, intra-step only, and intra+cross. Intra-step memory improves SR and stabilizes pose estimates across datasets; adding cross-step memory further improves long-horizon performance and achieves the best SR/RPE (0.78/0.11), showing that cross-step context provides complementary gains beyond intra-step stabilization.

\begin{table}[!t]
\small
\setlength{\abovecaptionskip}{0pt}
\setlength{\belowcaptionskip}{0pt}
\centering
\caption{\small \textbf{Impact of number of selected layers} included in memory bank on navigation performance of $\text{UniWM}$ (with $\mathcal{M}^{\text{intra}}_t$ \& $\mathcal{M}^{\text{cross}}_t$) on evaluation splits of four in-domain datasets. Best results for each metric are in bold.}
\label{tab:layernum_ablation_nav}
\resizebox{1\linewidth}{!}{%
\begin{tabular}{p{2cm}<{\centering}p{1cm}<{\centering}p{1cm}<{\centering}p{1cm}<{\centering}p{1cm}<{\centering}p{1cm}<{\centering}p{1cm}<{\centering}p{1cm}<{\centering}p{1cm}<{\centering}p{1cm}<{\centering}p{1cm}<{\centering}p{1cm}<{\centering}p{1cm}<{\centering}}
\toprule
\multirow{2}{*}{\textbf{Layer Num}} 
& \multicolumn{3}{c}{\textbf{Go Stanford}} 
& \multicolumn{3}{c}{\textbf{ReCon}} 
& \multicolumn{3}{c}{\textbf{SCAND}} 
& \multicolumn{3}{c}{\textbf{HuRoN}} \\
\cmidrule(lr){2-4} \cmidrule(lr){5-7} \cmidrule(lr){8-10} \cmidrule(lr){11-13}
& \textbf{SR $\uparrow$}& \textbf{ATE $\downarrow$}& \textbf{RPE $\downarrow$}& \textbf{SR $\uparrow$}& \textbf{ATE $\downarrow$}& \textbf{RPE $\downarrow$}& \textbf{SR $\uparrow$}& \textbf{ATE $\downarrow$}& \textbf{RPE $\downarrow$}& \textbf{SR $\uparrow$}& \textbf{ATE $\downarrow$}& \textbf{RPE $\downarrow$}\\
\midrule
1 
& 0.71 & 0.30 & 0.10
& 0.84 & 0.36 & 0.12
& 0.62 & 0.35 & 0.14
& 0.72 & 0.41 & 0.14 \\
\rowcolor{w_1!7} 3 
& 0.74 & 0.27 & 0.09
& 0.89 & 0.35 & 0.11
& 0.66 & 0.33 & 0.13
& 0.75 & 0.39 & 0.14 \\
\rowcolor{w_1!27} 5 
& \textbf{0.75} & \textbf{0.22} & \textbf{0.09} 
& \textbf{0.93} & \textbf{0.34} & \textbf{0.11} 
& 0.68 & 0.32 & \textbf{0.13} 
& \textbf{0.76} & \textbf{0.38} & \textbf{0.13} \\
\rowcolor{w_1!40} 7 
& 0.74 & 0.25 & 0.09
& 0.91 & 0.35 & 0.12
& \textbf{0.69} & \textbf{0.31} & 0.13
& 0.74 & 0.38 & 0.14 
\\
16 
& 0.61 & 0.46 & 0.23
& 0.70 & 0.58 & 0.26
& 0.52 & 0.49 & 0.24
& 0.57 & 0.55 & 0.22 
\\
32 
& 0.58 & 0.52 & 0.26
& 0.67 & 0.64 & 0.29
& 0.49 & 0.55 & 0.27
& 0.54 & 0.61 & 0.25 
\\
\bottomrule
\end{tabular}}
\end{table}

\begin{table}[!t]
\setlength{\abovecaptionskip}{0pt}
\setlength{\belowcaptionskip}{0pt}
\centering
\caption{\small \textbf{Comparison of navigation performance under different step strategies} across four datasets.}
\label{tab:step_nav_metrics}
\resizebox{1\linewidth}{!}{%
\begin{tabular}{lp{1cm}<{\centering}p{1cm}<{\centering}p{1cm}<{\centering}p{1cm}<{\centering}p{1cm}<{\centering}p{1cm}<{\centering}p{1cm}<{\centering}p{1cm}<{\centering}p{1cm}<{\centering}p{1cm}<{\centering}p{1cm}<{\centering}p{1cm}<{\centering}}
\toprule
\multirow{2}{*}{\textbf{Step Strategy}} 
& \multicolumn{3}{c}{\textbf{Go Stanford}} 
& \multicolumn{3}{c}{\textbf{ReCon}} 
& \multicolumn{3}{c}{\textbf{SCAND}} 
& \multicolumn{3}{c}{\textbf{HuRoN}} \\
\cmidrule(lr){2-4} \cmidrule(lr){5-7} \cmidrule(lr){8-10} \cmidrule(lr){11-13}
& \textbf{SR $\uparrow$}& \textbf{ATE $\downarrow$}& \textbf{RPE $\downarrow$}& \textbf{SR $\uparrow$}& \textbf{ATE $\downarrow$}& \textbf{RPE $\downarrow$}& \textbf{SR $\uparrow$}& \textbf{ATE $\downarrow$}& \textbf{RPE $\downarrow$}& \textbf{SR $\uparrow$}& \textbf{ATE $\downarrow$}& \textbf{RPE $\downarrow$}\\
\midrule
\textbf{Predict both} 
& 0.65 & 0.38 & 0.13
& 0.80 & 0.41 & 0.15
& 0.57 & 0.39 & 0.17
& 0.63 & 0.47 & 0.20 \\
\rowcolor{w_1!27} \textbf{Interleave }  
& \textbf{0.71} & \textbf{0.32} & \textbf{0.10} 
& \textbf{0.82} & \textbf{0.35} & \textbf{0.12} 
& \textbf{0.61} & \textbf{0.36} & \textbf{0.14} 
& \textbf{0.70} & \textbf{0.42} & \textbf{0.15} \\  
\bottomrule
\end{tabular}
}
\end{table}

\noindent \textbf{5.~How many layers should be included in the memory bank?} 

Table~\ref{tab:layernum_ablation_nav} varies the number of memory-augmented layers (with both $\mathcal{M}^{\text{intra}}_t$ and $\mathcal{M}^{\text{cross}}_t$ enabled). Moderate multi-depth integration (3--7 layers) steadily improves SR/ATE/RPE, with 5 layers offering the best trade-off. Dense integration (16--32 layers) degrades performance and increases compute and KV overhead; we therefore adopt 5 layers in all experiments.

\noindent \textbf{6.~Does goal conditioning affect generalization?} 

We retrain UniWM without the goal image in the visualization substep, keeping all other settings fixed. On unseen TartanDrive, removing goal conditioning reduces performance (SR 0.33 \vs 0.35, ATE 1.37 \vs 1.20, RPE 0.51 \vs 0.46), indicating that goal conditioning improves generalization rather than hindering it.

\noindent \textbf{7.~Why UniWM predicts action and observation at different substeps?}

Table~\ref{tab:step_nav_metrics} compares two strategies: \emph{predict both} (jointly outputting $\hat{a}_{t+1}$ and $\hat{o}_{t+1}$ in one forward pass) \vs \emph{interleave} (alternating planner and world-model substeps during both training and inference). Across all datasets, interleaving yields higher SR and lower ATE/RPE, empirically validating our design choice.

\begin{figure}[!t]
\setlength{\abovecaptionskip}{0pt}
\setlength{\belowcaptionskip}{0pt}
\begin{minipage}{0.44\linewidth}
\vspace{-0.2in}
\captionof{table}{\small \textbf{Zero-shot navigation performance} evaluated on TartanDrive (unseen) without finetuning.}
\label{tab:nav_metrics_unseen}
\centering
\resizebox{\linewidth}{!}{%
\begin{tabular}{lccc}
\toprule
\textbf{Method} & \textbf{SR $\uparrow$} & \textbf{ATE $\downarrow$} & \textbf{RPE $\downarrow$} \\
\midrule
GNM~\citep{shah2022gnm}        &  0.16  & 2.45   & 0.79    \\
VINT~\citep{shah2023vint}      &  0.13    &  2.38    &  0.79    \\
NoMaD~\citep{sridhar2024nomad} &  0.18    &  2.23    &  0.77    \\
Anole-7B~\citep{chern2024anole}   & 0.15     &  2.12    &  0.83    
\\
NWM~\citep{bar2025navigation}  & 0.27  & 1.61   & 0.62    
\\
\rowcolor{w_1!7} \textbf{\ours~(w/o $\mathcal{M}$)}         & 0.35  & 1.20   & 0.46    \\
\rowcolor{w_1!15} \textbf{\ours~(w/ only $\mathcal{M}^{\text{intra}}_t$)}    & 0.38  & 1.04   & 0.41    
\\
\rowcolor{w_1!27} \textbf{\ours~(w/ $\mathcal{M}^{\text{intra}}_t$ \& $\mathcal{M}^{\text{cross}}_t$)}  & \textbf{0.42}  & \textbf{0.95}   & \textbf{0.37}    
\\
\bottomrule
\end{tabular}}
\end{minipage}
\hfill
\begin{minipage}{0.54\linewidth}
\centering
\caption{\small \textbf{Qualitative Results} in unseen environments (from TartanDrive dataset) with UniWM. Red boxes denote ego-robot parts.}
\vspace{0.1cm}
\includegraphics[width=0.99\columnwidth]{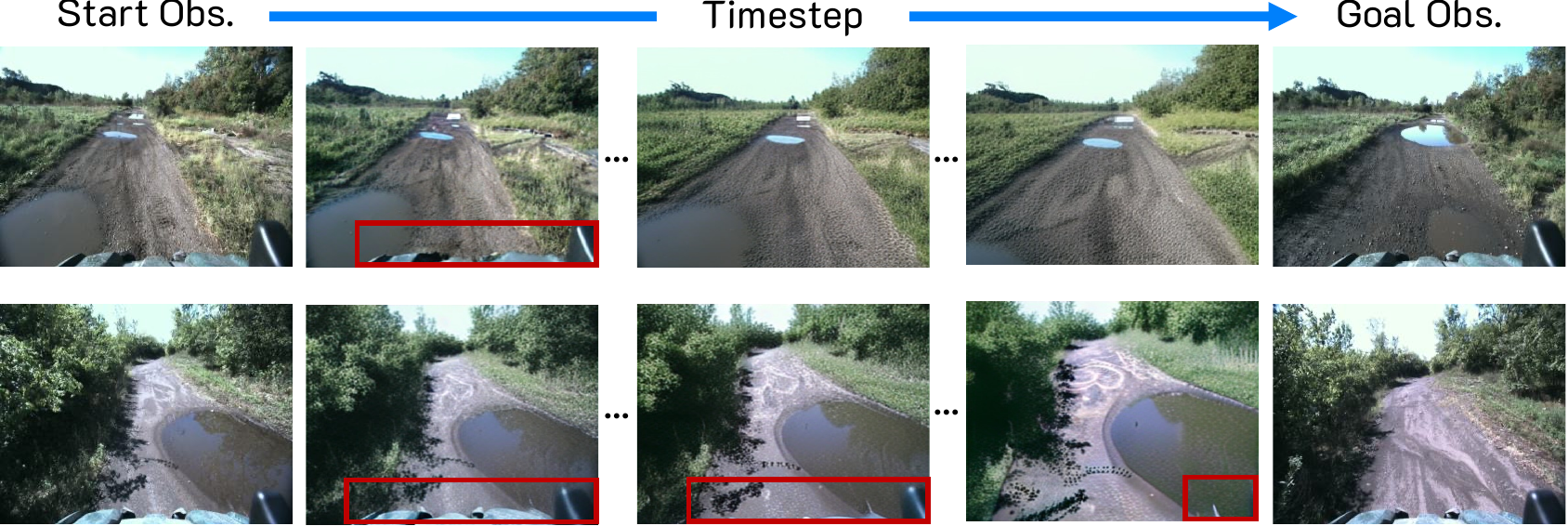}
\label{fig:tartan}
\end{minipage}
\end{figure}

\subsection{Generalization in Unseen Environments}
We evaluate zero-shot generalization on the unseen TartanDrive split (no fine-tuning) in Table~\ref{tab:nav_metrics_unseen} and Fig.~\ref{fig:tartan}. UniWM generalizes strongly: even without memory it improves SR and reduces pose errors. Adding $\mathcal{M}^{\text{intra}}_t$ stabilizes predictions, and further enabling $\mathcal{M}^{\text{cross}}_t$ improves long-horizon consistency, yielding best overall results. This confirms that UniWM transfers reliably to unseen environments.

\noindent \textbf{Error Cases \& Limitations.} TartanDrive observations occasionally contain visible ego-robot parts. As illustrated in Fig.~\ref{fig:tartan}, UniWM preserves these cues in the first-step prediction, but they may fade during rollouts. We attribute this to domain gap: the training datasets rarely include ego-robot regions, so the model treats them as background and effectively ``inpaints'' them away, leading to inconsistency with ground-truth frames in unseen settings.

\begin{figure}[!t]
\setlength{\abovecaptionskip}{0pt}
\setlength{\belowcaptionskip}{0pt}
\begin{minipage}{0.44\linewidth}
\vspace{-0.2in}
\captionof{table}{\small \textbf{Humanoid navigation performance assessment} with UniWM and other baselines trained from scratch on the 1X Humanoid Dataset.}
\label{tab:nav_metrics_humanoid}
\centering
\resizebox{\linewidth}{!}{%
\begin{tabular}{lccc}
\toprule
\textbf{Method} & \textbf{SR $\uparrow$} & \textbf{ATE $\downarrow$} & \textbf{RPE $\downarrow$} 
\\
\midrule
GNM~\citep{shah2022gnm}        &  0.41  & 0.82   & 0.29    \\

VINT~\citep{shah2023vint}      &  0.38    &  0.89    &  0.33    \\

NoMaD~\citep{sridhar2024nomad} &  0.44    &  0.76    &  0.27    \\

Anole-7B~\citep{chern2024anole}   & 0.20     &  1.95    &  0.73    \\

NWM~\citep{bar2025navigation}  & 0.51  & 0.57   & 0.24    \\

\rowcolor{w_1!7} \textbf{\ours~(w/o $\mathcal{M}$)}         & 0.78  & 0.31   & 0.15    \\

\rowcolor{w_1!15} \textbf{\ours~(w/ only $\mathcal{M}^{\text{intra}}_t$)}    & 0.79  & 0.29   & 0.14    \\

\rowcolor{w_1!27} \textbf{\ours~(w/ $\mathcal{M}^{\text{intra}}_t$ \& $\mathcal{M}^{\text{cross}}_t$)}  & \textbf{0.81}  & \textbf{0.28}   & \textbf{0.14}    
\\
\bottomrule
\end{tabular}}
\end{minipage}
\hfill
\begin{minipage}{0.52\linewidth}
\caption{\small \textbf{Qualitative Results} on 1X Humanoid Dataset with UniWM and NWM. The model generates egocentric observations consistent with 25-DoF joint-angle navigation commands.}
\centering
\includegraphics[width=\columnwidth]{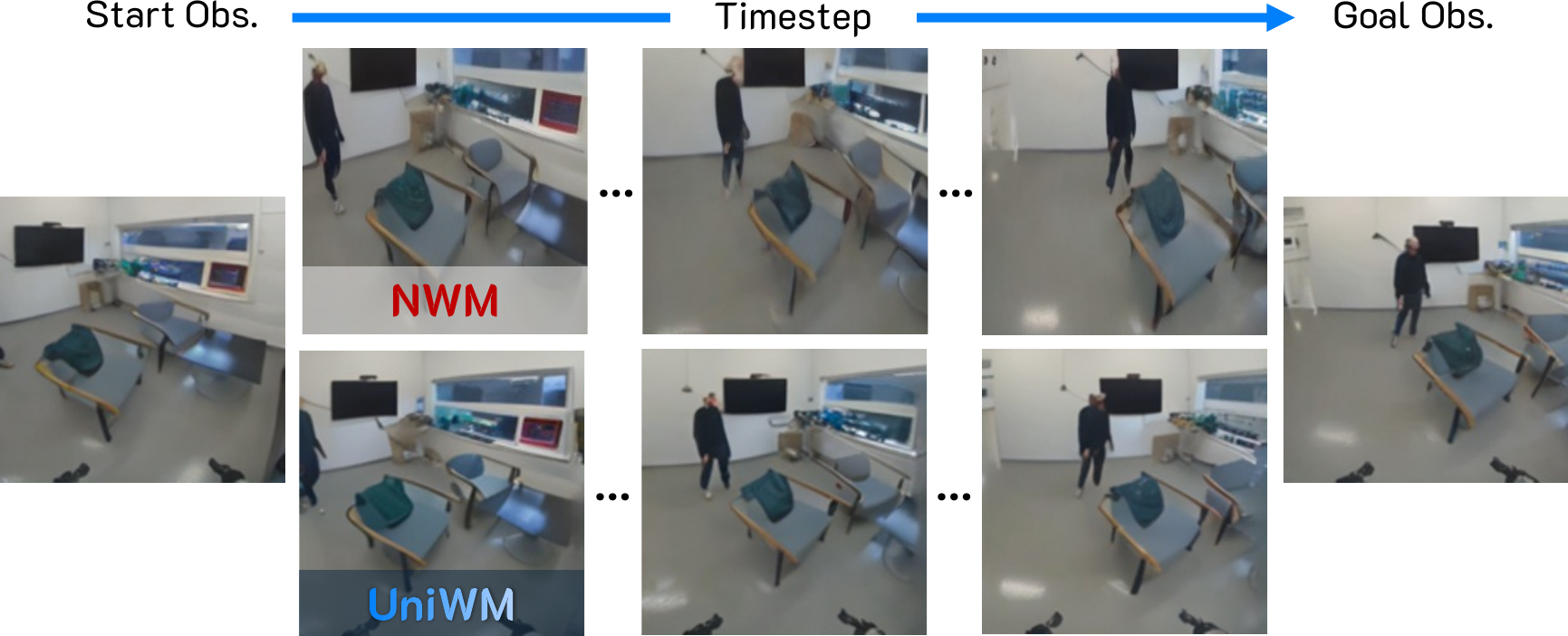}
\label{fig:humanoid}
\end{minipage}
\end{figure}

\subsection{Extension to Humanoid Navigation}
\label{sec:extension-humanoid}
To assess scalability to high-dimensional action spaces, we evaluate on the navigation subset of the 1X Humanoid Dataset~\citep{1X_Technologies_1X_World_Model_2024} with 25-DoF joint-angle commands. All baselines except Anole-7B (evaluated via direct prompting) are retrained from scratch under identical conditions. As shown in Table~\ref{tab:nav_metrics_humanoid}, UniWM consistently outperforms all baselines across SR, ATE, and RPE, and hierarchical memory provides further gains. Qualitative comparison in Fig.~\ref{fig:humanoid} shows that UniWM produces more spatially coherent rollouts than NWM, better preserving complex structures such as room layouts and ongoing human activities in the scene. UniWM also faithfully retains the humanoid robot's visible body parts (\eg, arms) in the egocentric view, which NWM tends to blur over time. 

%% file: sections/5_conclusion.tex
\section{Conclusion}
We presented \ours, a unified memory-augmented world model that couples visual imagination and navigation planning within a single multimodal autoregressive backbone. By jointly modeling perception, prediction, and control, UniWM mitigates state-action misalignment, while hierarchical memory fuses short-term observations with longer-range trajectory context to stabilize long-horizon rollouts. Across six benchmarks, including zero-shot evaluation on TartanDrive and 25-DoF humanoid navigation on the 1X Humanoid Dataset, UniWM achieves higher SR and lower ATE/RPE than strong baselines. Limitations include domain shift (\eg, ego-robot artifacts) and a fixed token budget; future work will explore adaptive token allocation, uncertainty-aware planning, and closed-loop deployment on real robots.

%% file: sections/X_appendix.tex
\section*{Appendix}
\addcontentsline{toc}{section}{Appendix}  

\setcounter{table}{0}  
\setcounter{figure}{0}

\setcounter{equation}{0}
\renewcommand{\thetable}{A\arabic{table}}
\renewcommand{\thefigure}{A\arabic{figure}}

\renewcommand{\theequation}{A\arabic{equation}}

This supplementary material provides additional details and results that complement the main paper. Sec.~\ref{appx. prompt} presents the detailed prompt design and examples for both action prediction and navigation visualization. Sec.~\ref{appx. pseudo-code} provides the pseudo-code for the hierarchical memory bank mechanism used during inference. Sec.~\ref{appx. eval metrics} describes the evaluation metric details, Sec.~\ref{appx. inference time} reports inference time comparisons, and Sec.~\ref{appx. more quality res} presents additional qualitative results.

\section{Method Details}

\subsection{Prompt Design and Examples}
\label{appx. prompt}
\begin{figure}[!b]
\setlength{\abovecaptionskip}{0pt}
\setlength{\belowcaptionskip}{0pt}
\newtcolorbox{fullpromptbox}[2]{
  enhanced,
  colback=#1!5,
  colframe=#1!50,
  fonttitle=\bfseries,
  title=#2,
  sharp corners,
  boxrule=0.8pt,
  width=\linewidth,
  lower separated=false,
  bottomtitle=0mm
}
\begin{fullpromptbox}{blue}{Action Prediction}
\textbf{\tcbox[colback=gray!20, colframe=gray!20, boxrule=0.1pt,top=0mm, bottom=0mm]{\quad \quad \quad \quad\quad\quad\quad\quad\quad\quad\quad\quad\quad\quad\quad\quad\quad\quad\quad\quad\quad\quad \textbf{Input} \quad \quad \quad \quad\quad\quad\quad\quad\quad\quad\quad\quad\quad\quad\quad\quad\quad\quad\quad\quad\quad\quad\quad}}

\textbf{Task:} Action Prediction

\textbf{Description:} Based on the current first-person observation, starting point observation and coordinate, goal point observation, predict the next action to take.

\textbf{Inputs:} Starting Pose: (-90.16528149, -187.79242581, 0.15229973)

Start observation:~~~\includegraphics[height=1.7cm]{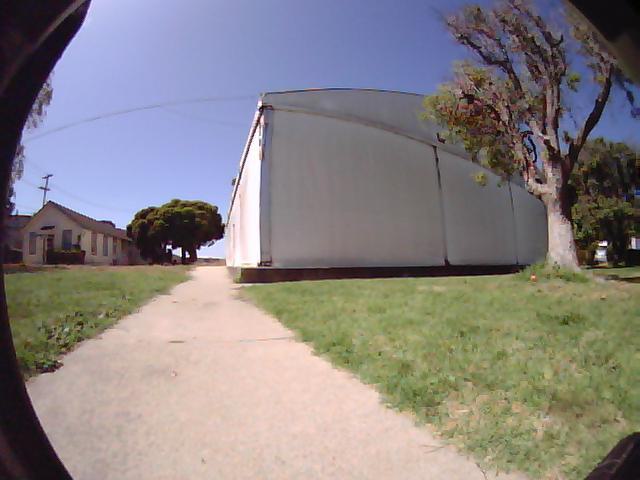}~~~Goal observation:~~~\includegraphics[height=1.7cm]{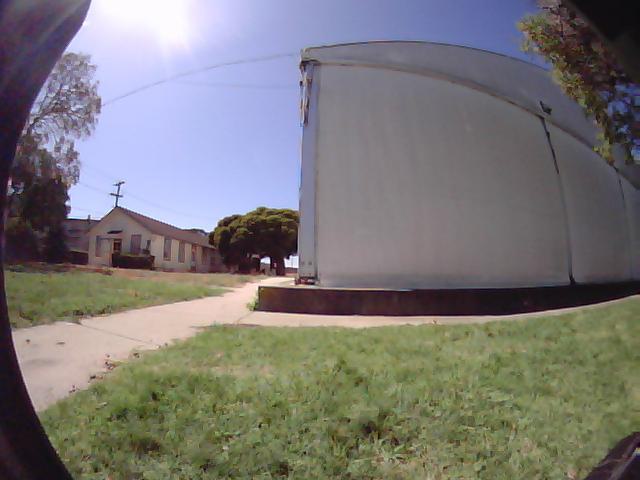}

Current observation:
~~~~~\includegraphics[height=1.7cm,width=2.265cm]{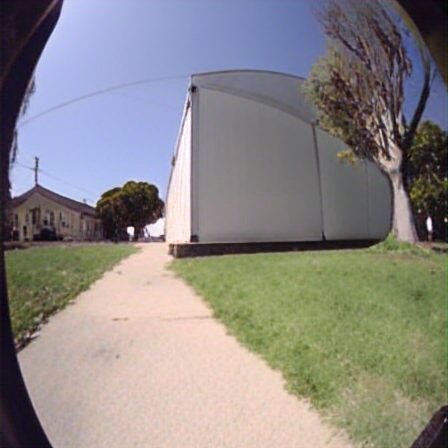}

\textbf{Action Format:}~The action can be the language command 'Stop', indicating the end of the trajectory. Alternatively, the action can be shifts composed of three components: - dx: displacement along the agent's facing direction), - dy: displacement perpendicular to the facing direction), - dyaw: change in heading angle (i.e., how much the agent rotates). All components are discretized into bin tokens: for example, - `dx pos bin 02`: dx = +0.02 meters, - `dy neg bin 23`: dy = -0.23 meters, - `dyaw pos bin 26`: counterclockwise rotation of +0.26 radians. If the agent reaches the goal or believes it has reached, it should predict 'Stop'.
-Output format: Move by dx: $<$dx$>$, dy: $<$dy$>$, dyaw: $<$dyaw$>$

\textbf{Goal:}~Predict the next action to approach the goal observation.

\textbf{\tcbox[colback=gray!20, colframe=gray!20, boxrule=0.1pt,top=0mm, bottom=0mm]{\quad \quad \quad \quad\quad\quad\quad\quad\quad\quad\quad\quad\quad\quad\quad\quad\quad\quad\quad\quad\quad\quad Response \quad \quad \quad \quad\quad\quad\quad\quad\quad\quad\quad\quad\quad\quad\quad\quad\quad\quad\quad\quad\quad}}
\textbf{Predicted Action:}

Move by dx:~$<$dx\_pos\_bin\_18$>$, dy:~$<$dy\_pos\_bin\_05$>$, dyaw:~$<$dyaw\_pos\_bin\_07$>$
\end{fullpromptbox}
\caption{\small Prompt design details and examples on action prediction (context size = 1).}
\label{fig:app_action}

\end{figure}

\begin{figure}[!t]
\setlength{\abovecaptionskip}{-2pt}
\setlength{\belowcaptionskip}{0pt}
\newtcolorbox{fullpromptbox}[3]{
  enhanced,
  colback=#1!5,
  colframe=#1!50,
  fonttitle=\bfseries,
  title=#2,
  sharp corners,
  boxrule=0.8pt,
  width=\linewidth,
  lower separated=false,
  bottomtitle=0mm
}
\begin{fullpromptbox}{teal}{Navigation Visualization}
\textbf{\tcbox[colback=gray!20, colframe=gray!20, boxrule=0.1pt,top=0mm, bottom=0mm]{\quad \quad \quad \quad\quad\quad\quad\quad\quad\quad\quad\quad\quad\quad\quad\quad\quad\quad\quad \quad\textbf{Input} \quad \quad \quad\quad\quad\quad\quad\quad\quad\quad\quad\quad\quad\quad\quad\quad\quad\quad\quad\quad}}

\textbf{Task:} Navigation Single Step Visualization

\textbf{Description:} Given the current first-person observation, predict the next first-person view observation after the agent executes a specified navigation action. To assist your prediction, you may refer to the start observation and pose (position: x, y and heading: yaw), as well as the goal and current observation.

\textbf{Inputs:} Next Action: Move by dx:~0.18, dy:~0.05, dyaw:~0.07

Starting Pose: (-90.16528149, -187.79242581, 0.15229973)

Start Observation:~~\includegraphics[height=1.7cm]{0.jpg}~~Goal Observation:~~\includegraphics[height=1.7cm]{9.jpg}

Current Observation:~~~~\includegraphics[height=1.7cm,width=2.265cm]{step_2_observation.png}

\textbf{Action Format:}~The action can be the language command 'Stop', indicating the end of the trajectory. Alternatively, the action can be shifts composed of three components: - dx: displacement along the agent's facing direction), - dy: displacement perpendicular to the facing direction), - dyaw: change in heading angle (i.e., how much the agent rotates). All components are discretized into bin tokens: for example, - `dx pos bin 02`: dx = +0.02 meters, - `dy neg bin 23`: dy = -0.23 meters, - `dyaw pos bin 26`: counterclockwise rotation of +0.26 radians.

\textbf{Spatial Interpretation:} - The magnitude of [dx, dy] reflects how far the agent moves in this step — larger values indicate greater positional shift, leading to larger visual changes. - dyaw controls the agent's rotation (change in heading). A positive dyaw indicates a left turn (counter-clockwise), while a negative dyaw indicates a right turn (clockwise).

\textbf{Goal:}~Predict the most likely next first-person observation, considering how the movement and rotation implied by `dx`, `dy`, and `dyaw` would affect what the agent sees next.

\textbf{\tcbox[colback=gray!20, colframe=gray!20, boxrule=0.1pt,top=0mm, bottom=0mm]{\quad \quad \quad \quad\quad\quad\quad\quad\quad\quad\quad\quad\quad\quad\quad\quad\quad\quad\quad\quad\quad\quad Response \quad \quad \quad \quad\quad\quad\quad\quad\quad\quad\quad\quad\quad\quad\quad\quad\quad\quad\quad\quad\quad}}
\textbf{Predicted observation:}~~~\includegraphics[height=1.7cm, width=2.265cm]{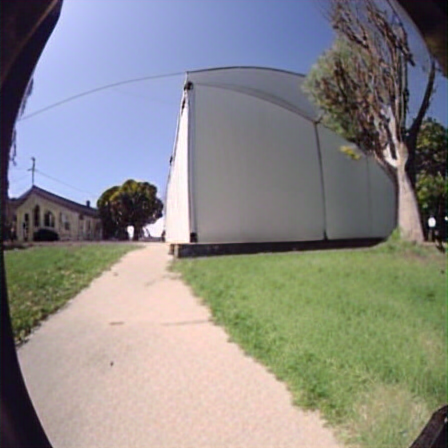}
\end{fullpromptbox}
\caption{\small Prompt design examples on navigation visualization (context size = 1).}
\label{fig:app_nv}

\end{figure}

We examine the detailed prompt formulation~\citep{liu2025grl,wang2025we} and response behaviors of two substeps: action prediction and navigation visualization in Figs.~\ref{fig:app_action} and \ref{fig:app_nv}. These examples illustrate how multimodal inputs guide both the navigation planner and the world model in visually grounded navigation.

\subsection{Pseudo-code for Hierarchical Memory Bank Mechanism}
\label{appx. pseudo-code}

\noindent Alg.~\ref{alg:uwm_mllm} details the inference process of UniWM, which systematically employs the hierarchical memory bank. The algorithm begins by initializing the intra-step memory $\mathcal{M}^{\text{intra}}_t$ and the persistent cross-step memory $\mathcal{M}^{\text{cross}}_t$ as empty sets (Line 9). It also defines a subset of decoder layers, $L_{\text{save}}$, from which Key-Value (KV) pairs will be extracted (Line 10). The main logic operates in a loop for each step $t$ from $1$ to $T$ (Line 11), divided into two substeps:

\noindent\textbf{Action Prediction.}~At the start of each step, the intra-step memory is cleared to prevent contamination from the previous state (Line 12). 
The \texttt{ExtractKV} function (Line 5, corresponding to Eq.~7) is invoked to extract KV pairs from the current observation $\hat{o}_{t-1}$, which are then stored in $\mathcal{M}^{\text{intra}}_t$ (Lines 13-14).
This new intra-step memory is then fused with the historical cross-step memory $\mathcal{M}^{\text{cross}}_t$ using the \texttt{Merge} function (Line 19), which encapsulates the spatio-temporal fusion logic from Eqs.~8,~9,~and 10. 
At the first step ($t=1$), when $\mathcal{M}^{\text{cross}}_t$ is empty, the fused memory $\tilde{\mathcal{M}}_t$ is simply the intra-step memory (Line 17).
Finally, the model predicts the action $\hat{a}_t$ using an enhanced attention mechanism conditioned on the fused memory $\tilde{\mathcal{M}}_t$, as described in Eq.~11 (Line 21).

\noindent\textbf{Navigation Visualization.}~Following action prediction, the model generates the next observation $\hat{o}_t$. This process reuses the same fused memory $\tilde{\mathcal{M}}_t$ from the action prediction substep, ensuring contextual consistency. The generation is conditioned on the prior state and the newly predicted action $\hat{a}_t$ (Line 23).

After both substeps, the intra-step memory $\mathcal{M}^{\text{intra}}_t$ is appended to the cross-step bank $\mathcal{M}^{\text{cross}}_t$, preserving context of current step for future predictions (Line 24).

This iterative process continues until the trajectory concludes, at which point the algorithm returns the complete sequences of predicted actions and observations (Line 26).

\begin{algorithm}[!t]
    \renewcommand{\algorithmicrequire}{\textbf{Input:}}
    \renewcommand{\algorithmicensure}{\textbf{Output:}}
\caption{Inference with Intra-step and Cross-step Memory Banks in UniWM}
    \label{alg:uwm_mllm}
    \begin{algorithmic}[1]
    \setstretch{0.7}
        \REQUIRE Start position $p_0$, start observation $o_s$, goal observation $o_g$; Decoder layers $L=\{l_0,\dots,l_{31}\}$
        \ENSURE Action sequence $A_T=\{\hat{a}_1,\dots,\hat{a}_T\}$, observation sequence $\mathcal{O}_T=\{\hat{o}_1,\dots,\hat{o}_T\}$

        \STATE \textbf{Definitions (helpers)}
        \STATE \hspace{0.6em}\texttt{ResetIntra}(): clear intra-step memory bank $\mathcal{M}^{\text{intra}}_t$
        \STATE \hspace{0.6em}\texttt{AppendIntra}($\{K^{(l)}_t,V^{(l)}_t\}_{l\in L_{\text{save}}}$): push layer-wise KV to $\mathcal{M}^{\text{intra}}_t$
        \STATE \hspace{0.6em}\texttt{AppendCross}($\mathcal{M}^{\text{intra}}_t$): push intra-step bank $\mathcal{M}^{\text{intra}}_t$ to cross-step bank $\mathcal{M}^{\text{cross}}$
        \STATE \hspace{0.6em}\texttt{ExtractKV}($\text{token seq.}$)$\!\rightarrow\!\{K^{(l)}_t,V^{(l)}_t\}_{l\in L_{\text{save}}}$: extract KV at selected layers (Eq.~7)
        \STATE \hspace{0.6em}\texttt{Merge}($\mathcal{M}^{\text{cross}}_t, \mathcal{M}^{\text{intra}}_t$)$\!\rightarrow\!\tilde{\mathcal{M}}_t$: memory fusion (Eqs.~8,~9,~and 10)
        \STATE \hspace{0.6em}\texttt{EnhanceAndDecode}(cond, $\mathcal{M}^{\text{intra}}_t$,$\tilde{\mathcal{M}}_t$)$\!\rightarrow\!$ predict with enhanced attention (Eq.~11)

        \STATE \textbf{Initialization}
        \STATE $\mathcal{M}^{\text{intra}}_t\leftarrow\varnothing$,\quad $\mathcal{M}^{\text{cross}}_t\leftarrow\varnothing$ \quad \COMMENT{cross-step memory is persistent across steps}
        \STATE $\hat{o}_0\leftarrow o_s$,\quad $L_{\text{save}} \leftarrow \{l_0,l_7,l_{15},l_{23},l_{31}\}$

        \FOR{$t=1$ \textbf{to} $T$}
            \STATE \texttt{ResetIntra}() \COMMENT{always reset intra-step memory at a new step}
            \STATE $\{K_{t}^{(l)}, V_{t}^{(l)}\} \leftarrow \texttt{ExtractKV}\big(p_0,o_s,o_g,\hat{o}_{t-1}\big)$
            \STATE \texttt{AppendIntra}\big($\{K_{t}^{(l)}, V_{t}^{(l)}\}_{l\in L_{\text{save}}}$\big)
            \STATE \textbf{Substep A: Action prediction at step $t$}
            \IF{$\mathcal{M}^{\text{cross}}_t=\varnothing$} 
                \STATE $\tilde{\mathcal{M}}_t \leftarrow \mathcal{M}^{\text{intra}}_t$ \quad \COMMENT{no cross memory at $t{=}1$}
            \ELSE
                \STATE $\tilde{\mathcal{M}}_t \leftarrow \texttt{Merge}\big(\mathcal{M}^{\text{cross}}_t, \mathcal{M}^{\text{intra}}_t)$
            \ENDIF
            \STATE $\hat{a}_t \leftarrow \texttt{EnhanceAndDecode}\big((p_0,o_s,o_g,\hat{o}_{t-1}),\, \tilde{\mathcal{M}}_t\big)$

            \STATE \textbf{Substep B: Navigation Visualization at step $t$}
            \STATE $\hat{o}_{t} \leftarrow \texttt{EnhanceAndDecode}\big((p_0,o_s,o_g,\hat{o}_{t-1}, \hat{a}_t),\,  \tilde{\mathcal{M}}_t\big)$
            \STATE \texttt{AppendCross}\big($\mathcal{M}^{\text{intra}}_t$\big) \quad             \COMMENT{Deposit intra-step memory to cross-step memory}
        \ENDFOR

        \STATE \textbf{return} $A_T=\{\hat{a}_1,\dots,\hat{a}_T\}$, $\mathcal{O}_T=\{\hat{o}_1,\dots,\hat{o}_T\}$
    \end{algorithmic}
\end{algorithm}

\section{Experiments and Results}

\subsection{Evaluation Metric Details}
\label{appx. eval metrics}
We evaluate overall system performance using two categories of metrics:

\noindent \textbf{Navigation Quality:}~For goal-conditioned visual navigation performance, the \textbf{Success Rate (SR)}~\citep{li2024human,dong2025ha} defines a trajectory as successful if its final distance $d$ to the goal is smaller than the agent’s average step size $\bar{s}$ (in meters). Formally, for trajectory $i$ among $N$ trajectories, with terminal estimate $\hat{p}^{(i)}_{T}$ and goal position $p^{(i)}_{g}$, SR is computed as:
$$
\text{SR}=\frac{1}{N}\sum_{i=1}^{N}\mathbf{1}\!\left[d\!\left(\hat{p}^{(i)}_{T},\,p^{(i)}_{g}\right)<\bar{s}\right],
$$
\textbf{Absolute Trajectory Error (ATE)} quantifies global trajectory accuracy by measuring the Euclidean distance between aligned points of the predicted and reference trajectories. \textbf{Relative Pose Error (RPE)} instead captures local consistency, computed as the deviation in relative motion between successive estimated and ground-truth poses~\citep{sturm2012evaluating,liu2026streamvlo}.

\noindent \textbf{(2) Visualization Quality:} For navigation visualization, visual predictions are evaluated with a combination of standard structural and perceptual measures, namely \textbf{SSIM}~\citep{wang2004ssim}, \textbf{PSNR}~\citep{hore2010image}, \textbf{LPIPS}~\citep{zhang2018unreasonable}, and \textbf{DreamSim}~\citep{fu2023dreamsim}. The latter two are deep perceptual metrics specifically designed to more closely approximate human judgments. To assess longer-horizon stability under rollout, we introduce four metrics: \textbf{SSIM$@n$}, \textbf{PSNR$@n$}, \textbf{LPIPS$@n$} and \textbf{DreamSim$@n$}. Standard one-step metrics compare ground-truth next frame $o_{t+1}$ with one-step prediction $\hat{o}^{(1)}_{t+1}$ obtained from ground truth current observation and action $(o_t,a_{t+1})$. For horizon $n$, we perform open-loop rollout that recursively feeds the model's predicted observations back as inputs while conditioning on ground-truth action sequence $a_{t+1:t+n+1}$: $\mathrm{SSIM}@n\!=\!\mathrm{SSIM}\!\big(o_{t+n},\,\hat{o}^{(n)}_{t+n}\big)$, where $\hat{o}^{(n)}_{t+n}$ is the observation prediction after $n$ rollouts, with PSNR$@n$, LPIPS$@n$ and DreamSim$@n$ defined analogously by replacing SSIM with the corresponding measure. 

\subsection{Inference Time}
\label{appx. inference time}
We provide a comparison of average inference time per trajectory together with navigation metrics across the four datasets (Go Stanford, ReCon, SCAND, HuRoN). As shown in Table \ref{tab:time}, NoMaD achieves fast inference but lacks imagination capability, which limits success rates in challenging cases. World-model-based approaches (NWM, Anole‑7B, UniWM) incur higher inference costs due to visual imagination. Importantly, UniWM runs substantially faster than its backbone Anole‑7B and NWM, while delivering markedly better navigation performance, demonstrating a favorable balance between efficiency and accuracy. Quantization to 4-bit \citep{frantar2022gptq} can potentially power UniWM up to an average of 16s per trajectory.

\begin{table}[t]
\setlength{\abovecaptionskip}{0pt}
\setlength{\belowcaptionskip}{0pt}
    \centering
    \caption{\textbf{Comparison of average inference time per trajectory} together with navigation metrics (SR, ATE, RPE) across the four datasets (Go Stanford, ReCon, SCAND, HuRoN).}
    \resizebox{1\linewidth}{!}{
    \begin{tabular}{l|cccc}
    \toprule
    Method    & Avg SR   & Avg ATE  & Avg RPE   &  Avg Inference Time per Trajectory (second) \\
    \midrule
         NoMaD \citep{sridhar2024nomad}& 0.48   & 0.79    & 0.26    &   0.6   \\
         NWM \citep{bar2025navigation}      &  0.55   &  0.63    &  0.23    &  114$\times$32 \\
         Anole-7B \citep{chern2024anole}&  0.27   & 1.80    &  0.73    &  654    \\
          UniWM (with Memory)      &  0.78    &  0.32     &   0.12     &  82        \\
           UniWM + Quant. 4-bit (with Memory)      &  -  &  -   &  -  &  16 (est. \citep{frantar2022gptq})          \\
           \bottomrule
    \end{tabular}}
    \label{tab:time}
\end{table}

\subsection{More Qualitative Results}
\label{appx. more quality res}
\begin{figure}[!t]
    \centering
    \includegraphics[width=0.98\linewidth]{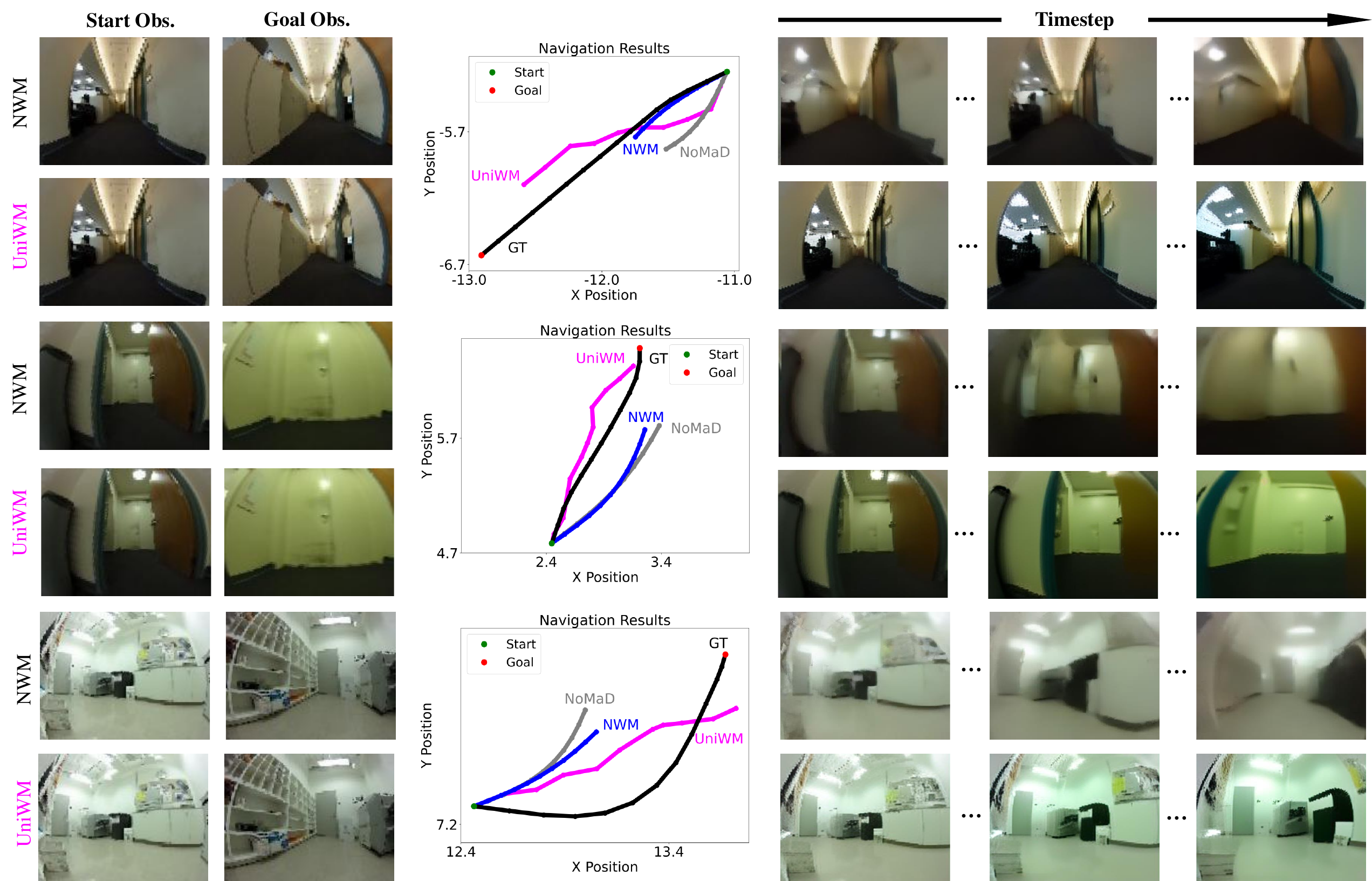}
    \vspace{-0.2cm}
    \caption{\small \textbf{Qualitative Comparisons} on Go Stanford across UniWM, NWM, and NoMaD. Central trajectory plots highlight the difference between predicted $A_T$ and GT.}
    \label{fig:visual2}
\end{figure}

\begin{figure}[!t]
    \centering
    \includegraphics[width=0.98\linewidth]{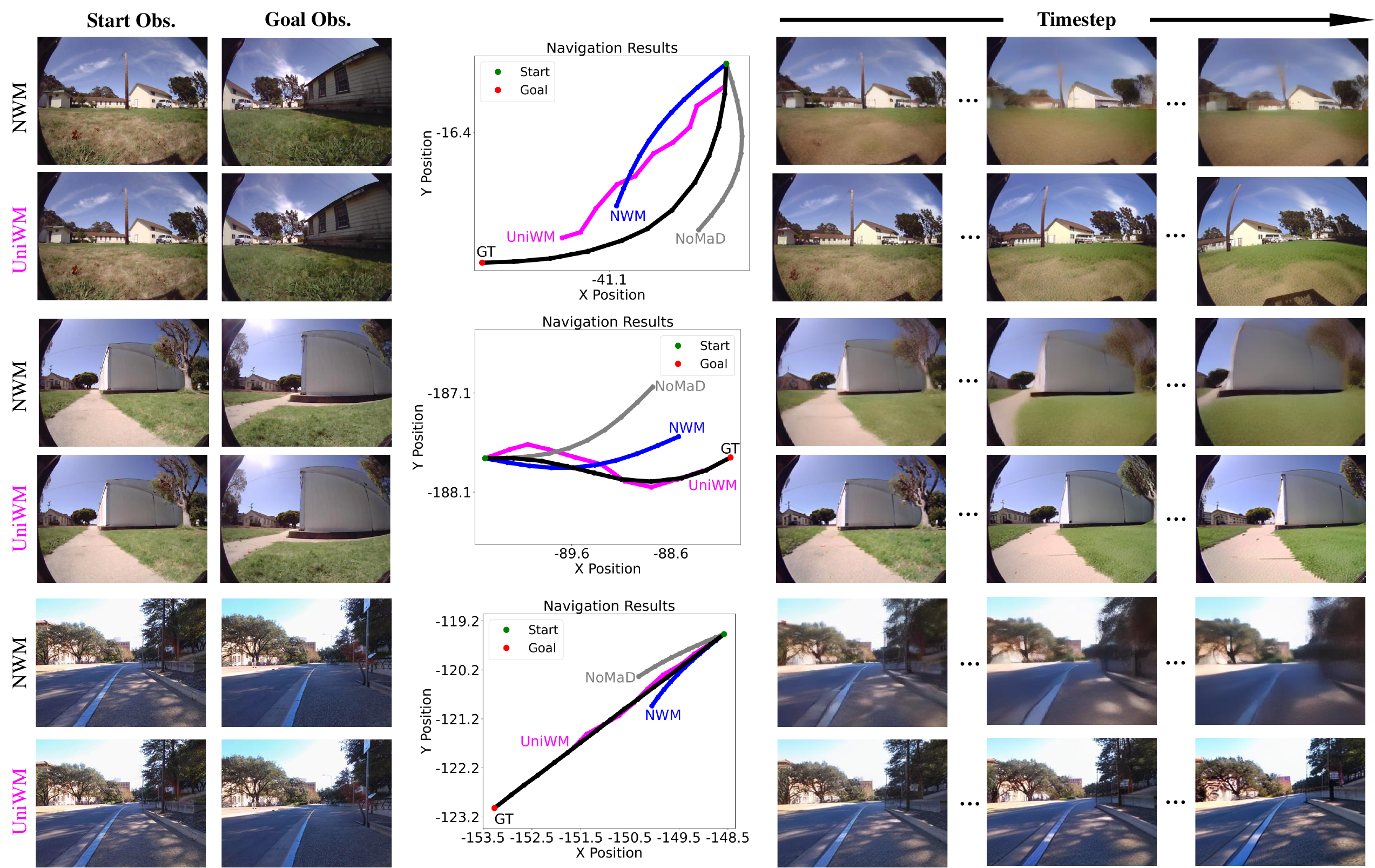}
    \vspace{-0.2cm}
    \caption{\small \textbf{Qualitative Comparisons} on ReCon and Scand across UniWM, NWM, and NoMaD. Central trajectory plots highlight the difference between predicted $A_T$ and GT.}
    \label{fig:visual3}
\end{figure}

\begin{figure}[!t]
    \centering
    \includegraphics[width=0.98\linewidth]{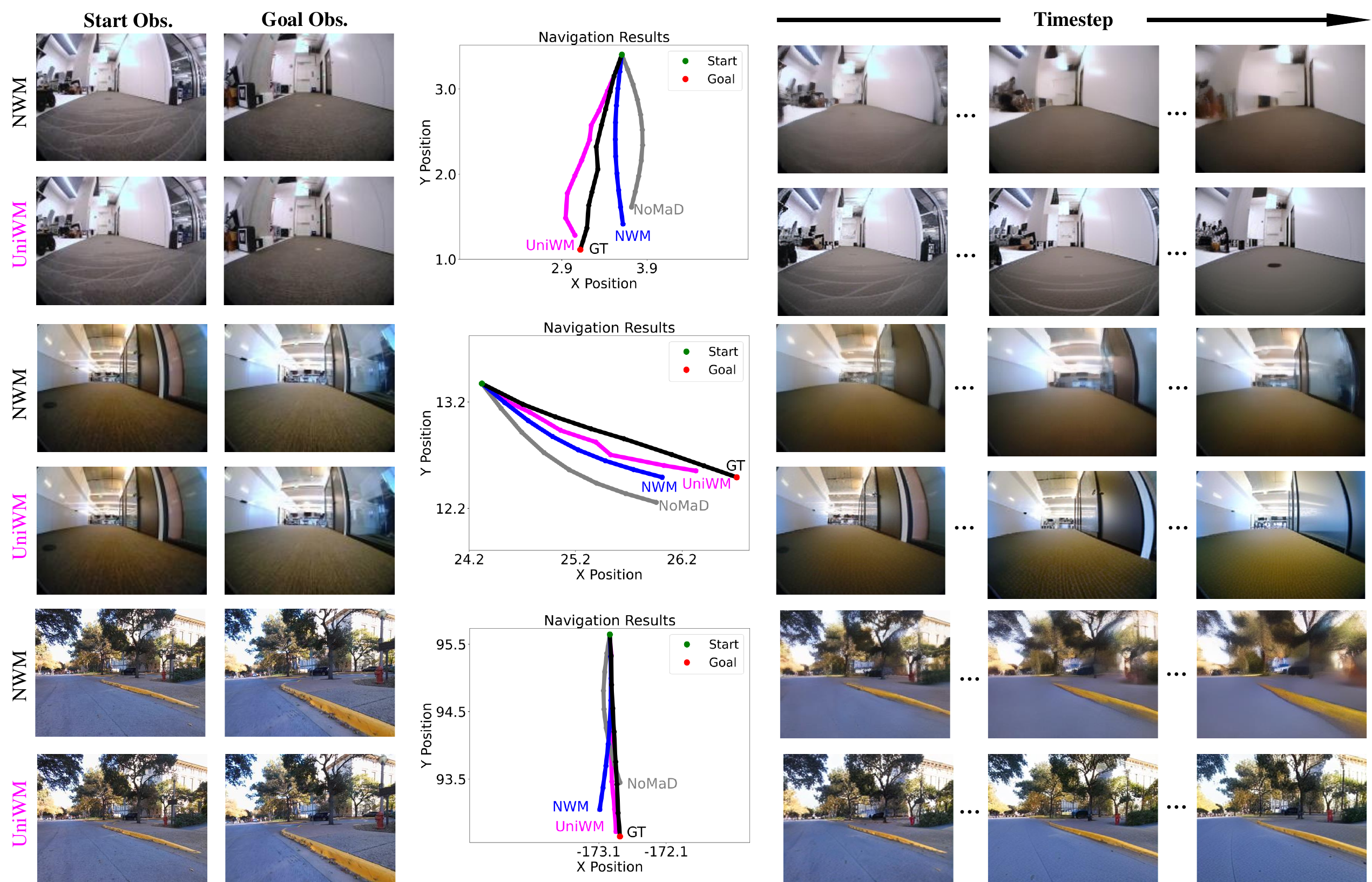}
    \vspace{-0.2cm}
    \caption{\small \textbf{Qualitative Comparisons} on HuRoN across UniWM, NWM, and NoMaD. Central trajectory plots highlight the difference between predicted $A_T$ and GT.}
    \label{fig:visual4}
\end{figure}

We provide more qualitative results in Figs.~\ref{fig:visual2}, \ref{fig:visual3} and \ref{fig:visual4}.

\section{Use of LLMs}
Large Language Models (LLMs) were only used to aid in sentence rephrasing and grammar checking of the manuscript.